\pdfoutput=1

\documentclass[11pt]{article}

\usepackage[]{EMNLP2023}
\usepackage{amsmath}
\usepackage{array}
\usepackage{bm}
\DeclareUnicodeCharacter{FFFD}{}
\usepackage{graphicx}   
\usepackage{enumitem}
\usepackage{lipsum}               
\usepackage{times}
\usepackage{seqsplit} 
\usepackage{amsmath} 
\usepackage{amssymb} 
\usepackage{tcolorbox}       
\usepackage{xcolor}          
\usepackage{graphicx}        
\usepackage{caption}         
\usepackage{calc} 
\usepackage{array}    
\usepackage{mathtools}
\usepackage{booktabs}    
\usepackage{multirow}    
\usepackage{makecell}    
\usepackage{geometry}    %
\usepackage{latexsym}
\usepackage{newunicodechar}
\usepackage{booktabs} 
\usepackage{tabularx} 
\usepackage{multirow} 
\usepackage{booktabs} 
\usepackage{soul}
\usepackage{tabularx} 
\usepackage{amsmath} 
\usepackage[T1]{fontenc}
\definecolor{lightblue}{rgb}{0.8,0.8,1}
\definecolor{lightpink}{rgb}{1,0.8,0.8}
\definecolor{lightyellow}{rgb}{1,1,0.8}

\usepackage[utf8]{inputenc}

\usepackage{microtype}

\usepackage{inconsolata}
\usepackage{hyperref}
\hypersetup{
    colorlinks=true,
    linkcolor=blue,
    citecolor=blue,
    filecolor=magenta,
    urlcolor=blue,
}

\usepackage{colortbl}
\definecolor{em}{gray}{0.9}
\newcommand{\cem}{\cellcolor{em}}

\title{Feature Extraction and Steering for Enhanced Chain-of-Thought Reasoning in Language Models}

\author{
Zihao Li$^{1}$\thanks{\,\,\, The first two authors contributed equally to this work.} \quad
Xu Wang$^{1}$\footnotemark[\value{footnote}] \quad
Yuzhe Yang$^2$ \quad
Ziyu Yao$^3$ \quad
Haoyi Xiong$^4$ \quad
Mengnan Du$^1$\\[4pt]
$^1$New Jersey Institute of Technology \quad 
$^2$University of California, Santa Barbara \quad\\
$^3$George Mason University \quad 
$^4$Microsoft \\[4pt]
\texttt{lizihao9885@gmail.com}, \texttt{mengnan.du@njit.edu}
}

\begin{document}
\maketitle

\begin{abstract}

Large Language Models (LLMs) demonstrate the ability to solve reasoning and mathematical problems using the Chain-of-Thought (CoT) technique. Expanding CoT length, as seen in models such as DeepSeek-R1, significantly enhances this reasoning for complex problems, but requires costly and high-quality long CoT data and fine-tuning. This work, inspired by the deep thinking paradigm of DeepSeek-R1, utilizes a steering technique to enhance the reasoning ability of an LLM without external datasets. Our method first employs Sparse Autoencoders (SAEs) to extract interpretable features from vanilla CoT. These features are then used to steer the LLM's internal states during generation. Recognizing that many LLMs do not have corresponding pre-trained SAEs, we further introduce a novel SAE-free steering algorithm, which directly computes steering directions from the residual activations of an LLM, obviating the need for an explicit SAE. Experimental results demonstrate that both our SAE-based and subsequent SAE-free steering algorithms significantly enhance the reasoning capabilities of LLMs.

\end{abstract}

\section{Introduction}


\begin{figure*}[htbp]
    \centering
    \includegraphics[width=\textwidth]{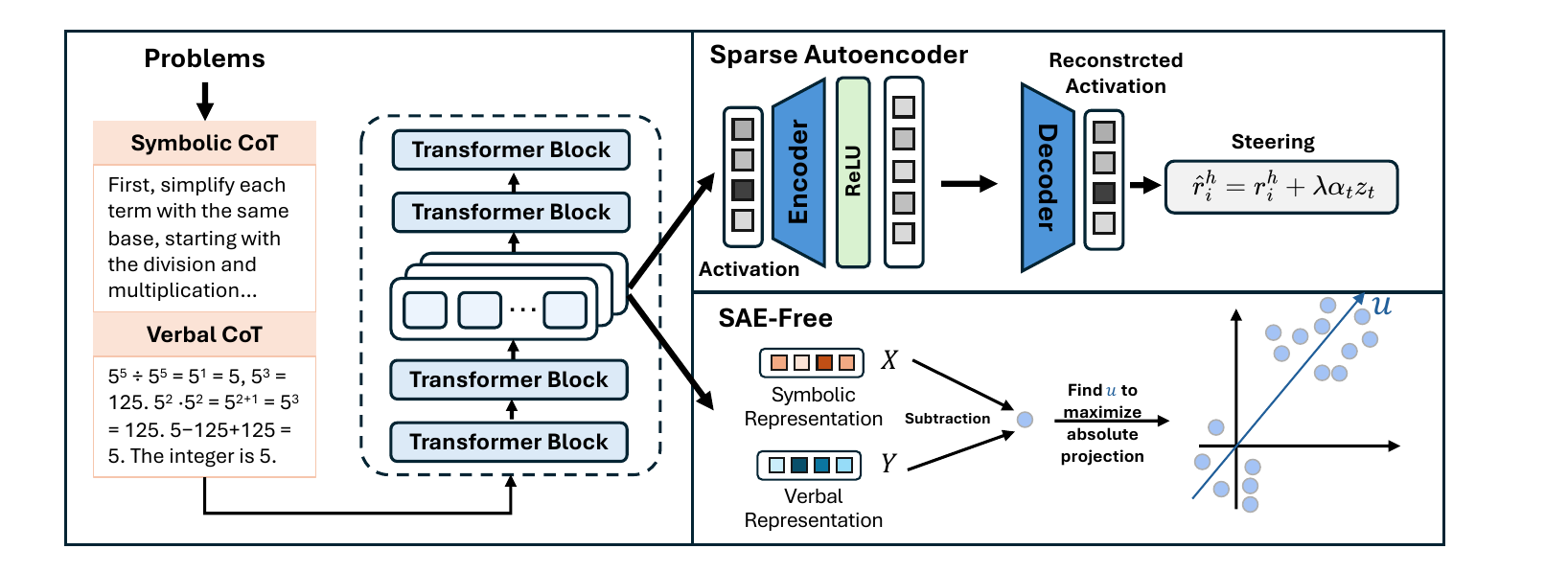} 
    \caption{Main framework of SAE-based steering and SAE-free algorithm.}
    \label{fig:main}
\end{figure*}

Large Language Models (LLMs), such as GPT-4~\cite{achiam2023gpt} and Claude3, have achieved remarkable success in natural language processing due to their large parameter size. They tackle logical reasoning and mathematical problems by employing CoT prompting~\cite{wei2023chainofthoughtpromptingelicitsreasoning} to enable step-by-step reasoning. Recent models such as DeepSeek-R1~\cite{guo2025deepseek}, OpenAI’s o1 series~\cite{jaech2024openai}, Qwen2.5~\cite{yang2024qwen2}, and Kimi1.5~\cite{team2025kimi} amplify this by expanding CoT length, incorporating more reasoning steps, self-correction, and backtracking compared to vanilla CoT. This enhancement is typically achieved through distillation (e.g., DeepSeek-R1-distill-LlaMa3-8B~\cite{guo2025deepseek}) or post-training~\cite{zeng2025simplerl,song2025fastcurl}. However, these methods for generating long CoT suffer from significant limitations. They need expensive, high-quality long CoT datasets which are difficult to obtain, and the process often overlooks the nuanced internal states of LLMs. Moreover, the verbosity of long CoT can introduce noisy features that do not contribute to, and may even detract from, the core reasoning process.

Concurrently, modulating LLM hidden states has proven effective for tasks such as knowledge editing~\cite{meng2022locating} and improving truthfulness~\cite{marks2023geometry,li2023inference}. Prior to our work, studies by \citet{tang2025unlocking}, \citet{sun2025thinkedit}, and \citet{hojer2025improving} have demonstrated that steering activations or editing weights can enhance the reasoning ability in smaller LLMs. For instance, \citet{tang2025unlocking} relies on contrasting representations (from long vs. short CoT) to derive steering vectors. While promising, these existing steering approaches may still depend on the generation of contrastive data or may not fully exploit the fine-grained feature distinctions within a single, standard reasoning trace. Furthermore, they typically do not offer a systematic way to disentangle task-relevant reasoning signals from stylistic or superficial features present in the CoT.


This paper introduces a novel framework that addresses these limitations by enabling precise steering of LLM residual activations using features extracted exclusively from readily available vanilla CoT, thereby bypassing the need for expensive long CoT datasets, contrastive examples, or model retraining. Inspired by the ``deep thinking'' mode of reasoning models, e.g., DeepSeek-R1, which suggests that crucial reasoning components are latently present even in standard CoT, our techniques aim to identify and amplify these capabilities.

First, the core of our proposal is a Sparse Autoencoder~(SAE)-based \emph{steering mechanism that uniquely targets reasoning-specific features}. We employ SAEs~\cite{cunningham2023sparse,shu2025surveysparseautoencodersinterpreting} to decompose the residual activations from vanilla CoT into interpretable features. To enhance feature selectivity and mitigate the influence of noise, we introduce a distinctive approach named \textbf{VS Decomposition}: the CoT is decomposed into a \textit{Verbal process} (natural language articulation) and a \textit{Symbolic process} (formal, rule-based logic), drawing inspiration from existing work such as~\citet{pan2023logic}. In addition, by computing the \textit{absolute difference} between SAE feature activations from these paired processes, we isolate features strongly indicative of either verbal or symbolic reasoning facet while effectively suppressing common, non-informative features (e.g., punctuation, sentence connectors). These distilled features then guide the steering of the LLM’s internal representations, promoting more focused and robust reasoning.



Additionally, we observe that even though SAE-based steering shows an incredible performance in enhancing LLMs' reasoning, there are still large amounts of LLMs that do not have associated SAEs, such as the Qwen series~\cite{yang2025qwen3technicalreport}. 
To address this issue, we design an \emph{SAE-free steering algorithm} to 
modulate the residual activation even without an associated SAE. The result shows that it achieves significant improvement in four datasets for mathematical reasoning.
The contribution of this paper is the following:
\begin{itemize}[leftmargin=*]\setlength\itemsep{-0.3em}
    \item We proposed a framework named \textit{VS Decomposition} to use SAE to extract features from vanilla CoT. It decomposes the reasoning process into two sub-processes: verbal process and symbolic process, enabling us to extract reasoning features while suppressing noise features.

    \item We proposed to steer the model by modulating the hidden activation employing the feature we obtained in the above step.

    \item We proposed an algorithm to steer LLMs without relying on SAE, which makes our steering method more applicable. The result shows that it achieves significant improvement in four mathematical reasoning datasets.
\end{itemize}

\section{SAE-based Steering}

\subsection{Preliminary on SAE}
Sparse Autoencoder (SAE) provides a method to interpret the hidden states of an LLM. It was trained to reconstruct the hidden representation. Given a hidden activation $h\in \mathbb{R}^{d_{\text{model}}}$, SAE encodes it into a sparse high-dimensional representation $\alpha \in \mathbb{R}^{d_{\text{SAE}}}$ and then decodes it to reconstruct $h$. The activation values in $\alpha$ are considered as features. 
These features can be explained by methods such as vocabulary projection~\cite{nostalgebraist2020interpreting} and visualization~\cite{sae_vis}, with the help of libraries such as neuronpedia.\footnote{\url{https://www.neuronpedia.org/}}

\begin{figure*}[htbp]
    \centering
    \includegraphics[width=\textwidth]{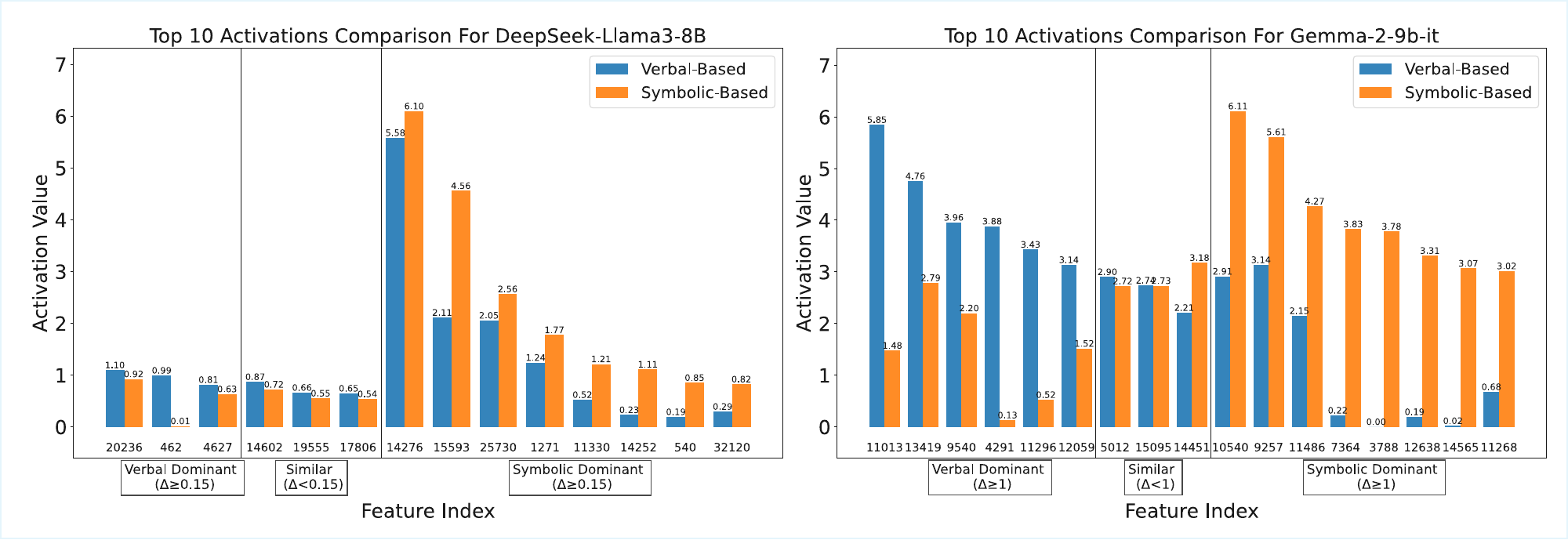} 
    \caption{Top 10 activations for DeepSeek-Llama3-8B and Gemma-2-9b-it. }
    \label{fig:top}
\end{figure*}

\subsection{Our Motivation}

Previous methods often obtained the correct features or directions by using the representation differences between long and short CoT~\cite{tang2025unlocking,sun2025thinkedit}. However, some flaws exist that we can not ignore. First, since it is costly to perform long CoT reasoning, long CoT datasets of high quality, such as Openthoughts~\cite{openthoughts}, are difficult to get. Second, compared with vanilla CoT, long CoT contains more reasoning features but also brings a large amount of noisy features such as ``punctuation marks'' or ``breaks of sentences''. Based on the above reasons, we want to find a method that can extract as many useful features as possible while avoiding noisy features. 

Recent work \cite{pan2023logic,quan2025peirce} thinks that CoT may imitate human reasoning processes due to a lack of a mechanism to guarantee the faithfulness of reasoning. In contrast, symbolic reasoning ensures inherent interpretability and verifiability by its reliance on well-defined logical principles that adhere rigorously to formal deductive systems.
 It illustrates the potential of unfaithful reasoning if one just relies on the common COT. 
Inspired by this claim, we proposed VS Decomposition which  decompose the CoT of reasoning problems into two sub-processes: \emph{\ul{Verbal Process}} and \emph{\ul{Symbolic Process}}. The former provides reasoning features, while the latter provides formal features. By integrating features from both of them, we can extract useful features while suppressing noisy features.


\subsection{VS Decomposition  Framework}
We proposed a novel framework for extracting features without relying on external datasets. It aims to maximize the extraction of reasoning-relevant features while simultaneously minimizing the effect of noisy features.

\vspace{3pt}\noindent\textbf{Separating Verbal and Symbolic Process.}
For reasoning tasks like GSM8K~\cite{cobbe2021training}, the first step involves generating two sub-processes: Verbal Reasoning Process and Symbolic Reasoning Process. For example, for the problem ``\textbf{Express $\bm{5^5 \div 5^4 - 5^3 + 5^2 \cdot 5^1}$ as an integer}''. It can be divided like this, Verbal Reasoning Process: \texttt{``First, simplify each term with the same base, starting with the division and multiplication. Next, perform the subtraction and addition operations to get the final integer result''} and the Symbolic Reasoning Process:\texttt{ ``$5^5 \div 5^4 =5^1 = 5$,$5^3 = 125$ $5^2 \cdot 5^1 = 5^{2+1} = 5^3 = 125$ $5 - 125 + 125 = 5$ The integer is 5.''}

\begin{table*}[htbp]
    \centering

    \renewcommand{\arraystretch}{1.1}  
    \setlength{\tabcolsep}{4pt}       
    \resizebox{\textwidth}{!}{
    \begin{tabular}{@{} l l *{5}{c} @{}}
        \toprule
        Model  & Version& GSM8K & MATH-L3\&L4 & MMLU-high & MathOAI \\
        \midrule
        \multirow{2}{*}{Llama3.1-8B-it}  & Original prompt& 60.33 &\textbf{21.46} & 24.81 &30.20  & \\
        &\cem{\textbf{SAE-based-steering (ours)}}&\cem\textbf{62.67}&\cem\textbf{21.46}&\cem\textbf{27.04}&\cem\textbf{32.80}&\\
        
        \midrule
        \multirow{2}{*}{ DeepSeek-Llama3-8B}  & Original prompt& 82.67 &78.11 & 72.96 &78.60  & \\
        &\cem{\textbf{SAE-based-steering (ours)}}&\cem\textbf{85.67}&\cem\textbf{79.83}&\cem\textbf{76.30}&\cem\textbf{83.40}&\\
        \midrule 
         \multirow{2}{*}{Gemma-2-9b-it}  & Original prompt& 88.00 &\textbf{51.07} & 51.11 &60.20  & \\
        &\cem{\textbf{SAE-based-steering (ours)}}&\cem\textbf{90.67}&\cem50.64&\cem\textbf{52.22}&\cem\textbf{62.60}&\\
        \bottomrule
    \end{tabular}}
    
    \vspace{-0.5em}  
    \caption{Overall accuracy (\%) of steering. }
    \label{tab:accuracy_comparison}
\end{table*}

\vspace{3pt}\noindent\textbf{SAE Feature Extraction.}
For each question $q$, we leverage the GPT4.1 API\footnote{\url{https://openai.com/index/gpt-4-1/}} to generate a Verbal Reasoning Process $x$ and a Symbolic Reasoning Process $y$ like this. 
Then we input both sub-processes into the LLM to extract residual activations $\bm{r}_i^{x}$ and $\bm{r}_j^{y}$ at layer $l$, where $i$ and $j$ represent the token position of $x$ and $y$. 
Then we employ SAE to extract features. In detail, if $\mathcal{X}$ and $\mathcal{Y}$ represent the set of token positions of $x$ and $y$, the features for the Verbal Reasoning Process and the Symbol Reasoning Process can be described as follows:
\begin{equation}
\begin{aligned}
    \bm{\alpha}^{x} &= \frac{1}{\|\mathcal{X}\|}\sum_{i\in \mathcal{X}} SAE(\bm{r}_i^{x}), \\
    \bm{\alpha}^{y}&= \frac{1}{\|\mathcal{Y}\|}\sum_{j\in \mathcal{Y}} SAE(\bm{r}_j^{y}),
\end{aligned}
\end{equation}
where $\bm{\alpha}^{x} \in \mathbb{R}^{d_{\text{SAE}}} $ and $\bm{\alpha}^{y}  \in \mathbb{R}^{d_{\text{SAE}}}$ are considered as the feature activation of Verbal Process and Symbolic Process at layer $l$.

\vspace{3pt}\noindent\textbf{Intergrating Features.}
To derive the integrated features denoted as $\bm{\alpha}$ from both reasoning processes, we can leverage the vector addition: $\alpha_t = \|\alpha_t^{x} +\alpha_t^{y}\|$, where $\alpha_t$ means the value of $\bm{\alpha}$ at index $t$, $\| \cdot \|$ means the absolute value symbol. It integrates information from both sub-processes. However, this approach suffers from noise amplification, particularly from task-irrelevant features (e.g., punctuation and sentence boundaries).
To address that, we observe that the activations of such task-irrelevant features are very close between both processes shown in Figure~\ref{fig:top}. We therefore propose to compute the \emph{absolute difference} between activations, effectively suppressing noisy features while preserving relevant features. Specifically, the final CoT feature activation for question $q$ can be described as $\alpha_t= \|\alpha^{x}_t-\alpha^{y}_t\|$.


To better explain the motivation of our method. We derived the top-10 feature activations of the Verbal Reasoning Process and the Symbolic Reasoning Process by several representative examples. The result is shown as Figure \ref{fig:top}. For each sub-figure, the feature indexs are categorized into three groups, the left group represents features predominantly associated with verbal featurs ($\alpha_t^x \gg \alpha_t^y$), features in the middle group indicates balanced features between both processes, the right group represents features predominantly associated with symbolic features ($\alpha_t^x \ll \alpha_t^y$). We can find that for \texttt{DeepSeek-Llama3-8B}, features in the middle group, such as 19555 and 14602 mean ``punctuation marks, specifically commas'' which is unrelated to reasoning; features index 462 in the left group present big difference between two processes, it means ``relationship statements involving variables''; features in the right group like 15593 means ``numeric expressions and mathematical components''. For \texttt{Gemma-2-9b-it}, the phenomenon is almost the same as the former: feature 5012 and feature 14451 in the middle group are unrelated to reasoning.  These extracted results are the same as the assumption we made at first, which means subtraction can better erase noisy features while integrating useful information.

In practice, we randomly sample $N = 100$  samples:$\{q_p\}_{p=1}^{N}$ from the MATHOAI dataset\footnote{\url{https://huggingface.co/\\datasets/heya5/math_oai}} to generate the Verbal Reasoning Process $ \{x_p\}_{p=1}^{N}$ and the  Symbolic Reasoning Process $\{y_p\}_{p=1}^{N}$. The reason for selecting this dataset is that we want to ensure that the difficulty of the problem is not too complicated to avoid generating long CoTs, and not too simple to avoid generating indistinguishable CoTs for both processes. The final feature activation $\bm{\alpha}_t$ derived from $\{q_k\}_{l=1}^{N}$ can be described as follows:
\begin{gather}
\alpha_t= \frac{1}{N}\sum_{p=1}^N\| \alpha^{x_p}_t-\alpha^{y_p}_t\|
\label{eq:goal}
\end{gather}

\subsection{Steering}
After obtaining $\{\alpha_t\}_{t=1}^{d_{\text{SAE}}}$ from the last step, we sort it in descending order to get the top-k important SAE features.  We leverage a certain feature among them for steering, which modulates the residual activation of LLMs to control their behaviors. 
For a given feature indexed by $t$, inspired by~\citet{templeton2024scaling, farrell2024applying, galichin2025have}, we proposed to use the activation value of 
$\bm{\alpha}$ to modify the residual activation $\bm{r}^h$ as follows:
\begin{gather}
    \bm{\hat{r}}_{i}^{h}= \bm{r}_{i}^{h} + \lambda  \alpha_t \bm{z_t}, i\in \mathcal{H} 
\end{gather}

The modification is made in the layer $l$ according to the SAE version
we choose, $\lambda$ is the hyperparameters of strength, $\bm{z}_t = \bm{W}_{\text{dec},t} = \bm{W}_{\text{dec}}[t,:]$.






    

\section{Experiments of SAE-based Steering}
\begin{table*}[htbp]
    \centering
    \renewcommand{\arraystretch}{1.1}  
    \setlength{\tabcolsep}{4pt}       
    \resizebox{\textwidth}{!}{
    \begin{tabular}{@{} l l *{4}{c} @{}}
        \toprule
        Model & Version & GSM8K & MATH-L3\&L4 & MMLU-high & MathOAI \\
        \midrule
        \multirow{4}{*}{DeepSeek-Llama3-8B}  
        & Original prompt & 82.67 & 78.11 & 72.96& 78.60 \\
        & BoostStep~\cite{zhang2025booststep}&79.33 & 77.25& 71.11 & 78.80\\
        & Mathneuro~\cite{christ2024math}&84.67&78.54&68.89&76.00 \\
        & \cem{\textbf{SAE-free-steering (Ours)}} & \cem\textbf{85.67} & \cem\textbf{82.83} & \cem\textbf{78.15} & \cem\textbf{83.00} \\
        \midrule
        \multirow{4}{*}{DeepSeek-qwen-1.5B} 
        & Original prompt & 70.00 & 65.24 & 65.56 & 74.20\\
        & BoostStep~\cite{zhang2025booststep}&67.33 &66.09 & 63.70 & 73.40\\
        & Mathneuro~\cite{christ2024math} &72.66&66.09&63.33&73.00\\
        & \cem{\textbf{SAE-free-steering (Ours)}} & \cem\textbf{74.00} & \cem\textbf{69.10} & \cem\textbf{66.67} & \cem\textbf{77.00} \\
        \bottomrule
    \end{tabular}}
    
    \vspace{-0.5em}  
    \caption{Performance comparison (accuracy \%) between original LLMs and SAE-Free steering versions.}
    \label{tab:sae_free}
\end{table*}

\begin{figure*}[htbp]
    \centering
    \includegraphics[width=\textwidth]{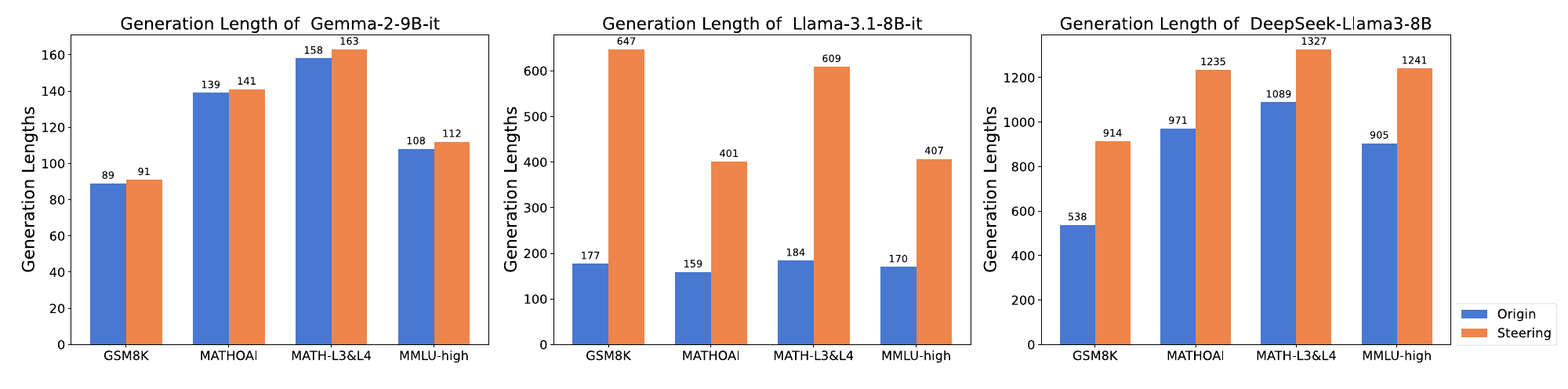} 
    \caption{The generation length variation of the questions with the top 25\% shortest answers generated by the original LLM.}
    \label{fig:length}
\end{figure*}
\subsection{Experimental Settings}
\noindent\textbf{Datasets.}\,\,
We evaluate the effect of our steering method on four mathematical benchmarks, including GSM8K~\cite{cobbe2021training}, 
MMLU-high school mathematics (MMLU-high)~\cite{hendryckstest2021}, MATH-500~\cite{lightman2023let}\footnote{\url{https://huggingface.co/datasets/HuggingFaceH4/MATH-500}} and MATHOAI\footnote{\url{https://huggingface.co/datasets/heya5/math_oai}}. For the MATH-500 dataset, we select MATH-level3 and MATH-level4 for testing. For the GSM8K dataset, to find the best parameters and features more efficiently, we conduct our experiment on a subset that contains 300 randomly selected samples instead of on the entire dataset. The baseline results are detailedly shown in appendix Table \ref{tab:baseline_gsm8k}. It means the results of subsets are roughly the same as the results on the entire dataset, thus ensuring that the randomly selected samples can be used to explore the effect of the steering experiment.

\vspace{3pt}
\noindent\textbf{LLM Models and SAEs}\,\,
Because we want to evaluate LLMs that have been proven to have excellent thinking ability, but at the same time, our method is based on pre-trained SAE, so we choose the \texttt{\seqsplit{DeepSeek-Llama3-8B}}~\cite{guo2025deepseek} and its base version  \texttt{\seqsplit{Llama3.1-8B-it}}~\cite{grattafiori2024llama}. Besides, we also choose \texttt{\seqsplit{Gemma-2-9b-it}}~\cite{team2024gemma} as evaluation LLM.
For SAEs, we leverage \texttt{\seqsplit{llama\_scope\_r1\_distill(l15r\_800m\_openr1\_math)}}\footnote{\url{https://huggingface.co/fnlp/Llama-Scope-R1-Distill}} to extract features from \texttt{\seqsplit{DeepSeek--Llama3-8B}} and \texttt{Llama3.1-8B-it}, \texttt{\seqsplit{gemma-scope-9b-it-res(layer\_9/width\_16k/canonical)}}\footnote{\url{https://huggingface.co/google/gemma-scope-9b-it-res}} to extract features from \texttt{Gemma-2-9b-it}.

\vspace{3pt}
\noindent\textbf{Implementation Details.}\,\,
For the generation setting, we employed a repetition penalty of 1.2 while setting distinct maximum sequence lengths for each dataset: 3000 tokens for GSM8K, 2500 for MATH-500, 3000 for MMLU-High, and 3500 for MATHOAI. 
For evaluation, we leverage LLM as a judge~\cite{phan2025humanity} to evaluate the accuracy of predictions. We utilize the GPT4.1 API as the judge. More details can be found in Appendix~\ref{app:steer}.

\subsection{Experimental Results Analysis}

\noindent\textbf{Comparing with Baselines.}
Table \ref{tab:accuracy_comparison} reports the overall accuracy results of four mathematical reasoning datasets. In addition to MATH-L3\&L4, there is a consistent improvement in the other three datasets. Especially for \texttt{DeepSeek-Llama3-8B}, steering improves the accuracy by 3.34\% and 4.80\% in MMLU-high and MathOAI. There is no significant improvement on MATHL3\&L4 for three LLMs, only \texttt{DeepSeek-Llama3-8B} achieves 1.72\% improvement, which means that SAE steering may play a better role in the reasoning model compared to the instruction model.
Through further manual reviewing, we found that for non-reasoning LLM \texttt{Llama3.1-8B-it}, too long generation length will lead the LLM to generate some irrelevant information, steering may improve this phenomenon by generating some thinking process related to the steering feature.

\vspace{3pt}
\noindent\textbf{Steering Effects on Reasoning Length.}
We also compared the length variation of the questions with the top 25\% shortest answers generated by the original LLM before and after steering.
The result is shown in Figure~\ref{fig:length}. Our experimental results indicate a steady enhancement in reasoning length for all evaluated datasets, with \texttt{Llama3.1-8B-it} exhibiting a significant increase. After observation, we found that the answers after steering will increase the number of reasoning paths and the depth of reasoning.
\begin{figure}[t]
    \centering
    \includegraphics[width=\columnwidth]{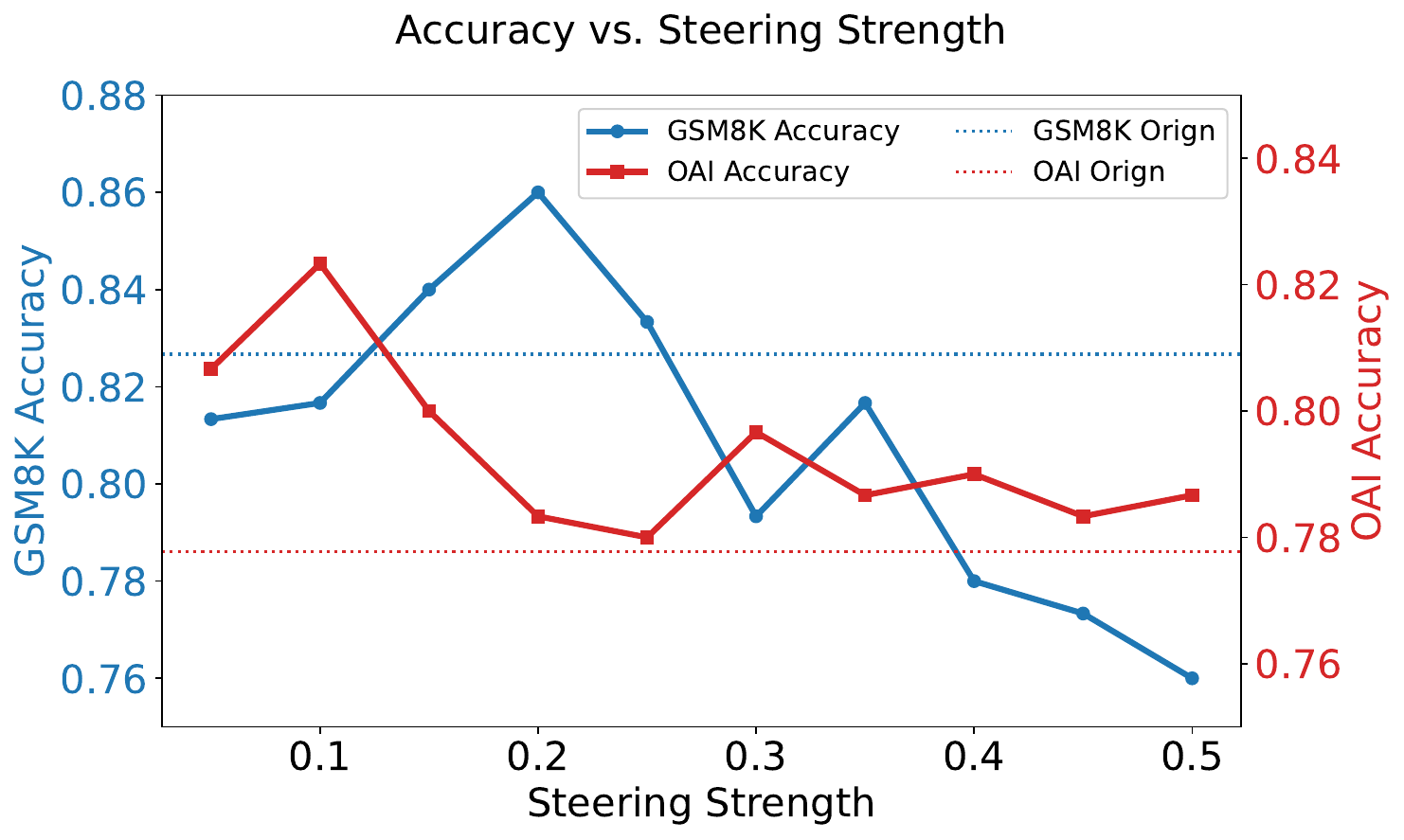} 
    \caption{Relationship between strength $\lambda$ and accuracy in GSM8K and MATHOAI.}
    \label{fig:curve}
\end{figure}

\vspace{3pt}
\noindent\textbf{Relationship between Strength and Accuracy.}
Taking \texttt{DeepSeek-Llama3-8B} as an example, we plotted the relationship between steering strength and accuracy. From Figure \ref{fig:curve} we can see that a higher strength does not always improve the accuracy of the model's response. There is an optimal strength, and exceeding it will cause the model's performance to decline. Manually checking the cases found that when the strength is too large, the model will generate many meaningless answers. This may be because excessively large strength will destroy the internal representation of the model.

\section{SAE-Free Steering}
\subsection{Motivation}
Feature-based steering has been proven to be useful in various fields of natural language processing, such as machine unlearning~\cite{muhamed2025saes,farrell2024applying}, improving truthfulness~\cite{marks2023geometry}, and interpretability~\cite{he2025saif,kharlapenkoscaling}. Although there are some trained SAEs released to help interpret hidden activations of LLMs, such as Gemma Scope~\cite{lieberum2024gemma} and Llama Scope~\cite{he2024llama}, there are still large amounts of LLMs that do not have matching SAEs to interpret them, such as the Qwen series. However, training an SAE is time-consuming and requires a lot of cost, so it is obviously unlikely to train an SAE for each layer of each LLM.

Based on the above reason, our method is hard to generalize to the scenarios in which there is no corresponding SAE. To solve this problem, we proposed an algorithm to generalize such an SAE-based steering method to LLMs without the corresponding SAE. 
Intuitively, we will start by assuming the availability of an SAE for a given LLM, and then derive the calculation to see if the SAE components can be approximated by others.

\begin{figure*}[t]
    \centering
    \includegraphics[width=\textwidth]{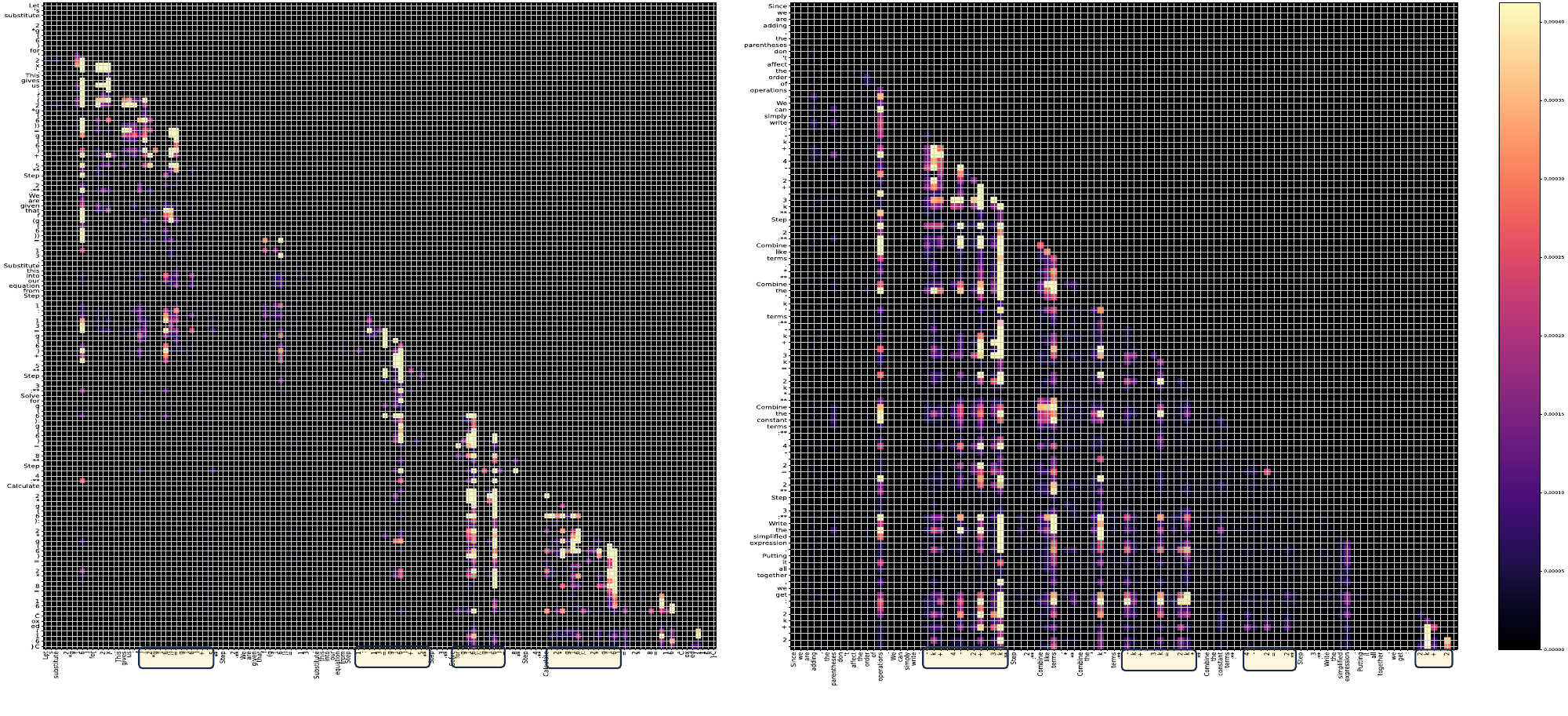} 
    \caption{Attention mapping of \texttt{\seqsplit{DeepSeek-Llama3-8B}} and \texttt{\seqsplit{{DeepSeek-qwen-1.5B}}}.} 
    \label{fig:map}
\end{figure*}

\subsection{Framework}

\noindent\textbf{Feature activation for virtual SAE.} SAE is trained to minimize the reconstruction loss for the hidden activation $\bm{h}$ of the layer $l$. Therefore, $\bm{h}$ can be decomposed as the weighted sum of  feature vectors$\{\bm{z}_t\}_{t=1}^{d_{SAE}}$ and error term $\bm{\epsilon}$, and $\alpha_t^{h}$  denotes the activation value of feature $t$ for $\bm{h}$ at layer $l$.
\begin{equation}
    \bm{h}  =\sum_{t=1}^{d_{\text{SAE}}} \alpha_t^{h} \bm{z}_t + \bm{b}+ \bm{\epsilon}
    \label{eq:sae}
\end{equation}

Now we want to steer the model even though there is no corresponding SAE model.
In our method, there are two types of CoT: Verbal Process $x$ contains $m$ tokens and Symbolic Process $y$ contains $n$ tokens.
We employ the mean representation in the layer $l$ to represent $x$ and $y$: $\bm{x} = \frac{1}{m}\sum_{i=1}^m \bm{r}_{i}^{x}$ ,$\bm{y} = \frac{1}{n}\sum_{j=1}^n \bm{r}_{j}^{y}$. Therefore, the $l$ layer's residual activation of $x$ and $y$ can be written as follows according to equation \eqref{eq:sae} :

\begin{subequations}
\label{eq:combined}
\begin{align}
\bm{x} &=  \frac{1}{m}\sum_{i=1}^m \bm{r}_{i}^{\bm{x}}
    = \frac{1}{m}  \sum_{i=1}^m ( \sum_{t=1}^{d_{\text{SAE}}}\alpha_t^{\bm{r_{i}^{x}}}\bm{z}_t+\bm{b} +\bm{\epsilon} ) \nonumber \\
    & = \sum_{t=1}^{d_{\text{SAE}}} \alpha_t^{\bm{x}} \bm{z_t} +\bm{b} + \frac{1}{m}\sum^m \bm{\epsilon}
    \label{eq:x_part} \\
\bm{y} &= \frac{1}{n} \sum_{j=1}^n \bm{r}_{j}^{\bm{y}}
    = \frac{1}{n}  \sum_{j=1}^n ( \sum_{t=1}^{d_{\text{SAE}}}\alpha_t^{\bm{r}_{j}^{\bm{y}}}\bm{z}_t+\bm{b} +\bm{\epsilon} ) \nonumber \\
    & = \sum_{t=1}^{d_{\text{SAE}}} \alpha_t^{\bm{y}} \bm{z}_t +\bm{b} + \frac{1}{n}\sum^n \bm{\epsilon}
    \label{eq:y_part}
\end{align}
\end{subequations}




Here $\alpha_t^{\bm{x}} = \frac{1}{m}\sum_{i=1}^m\alpha_t^{\bm{r}_{i}^{\bm{x}}}$ and $\alpha_t^{\bm{y}} = \frac{1}{n}\sum_{j=1}^n\alpha_t^{\bm{r}_{j}^{\bm{y}}}$ represent the mean activation of all tokens in $\bm{x}$ and $\bm{y}$ for feature $\alpha_t$ at layer $L$. 
Back to our SAE-based steering, to obtain the activation for each feature $\alpha_t$, we computed $\| \alpha_t^{\bm{x}}- \alpha_t^{\bm{y}} \|$. It can be derived as follows:
\begin{equation}
    \begin{aligned}
        \scriptstyle
        \bm{x} - \bm{y} &= \sum_{t=1}^{d_{\text{SAE}}} \alpha_t^{\bm{x}} \bm{z}_t +\bm{b} + \frac{1}{m}\sum^m \bm{\epsilon} \\ 
        &\quad - \sum_{t=1}^{d_{\text{SAE}}} \alpha_t^{\bm{y}} \bm{z}_t -\bm{b} - \frac{1}{n}\sum^n \bm{\epsilon} \\
        &= \sum_{t=1}^{d_{\text{SAE}}} \bm{z}_t \left( \alpha_t^{\bm{x}} - \alpha_t^{\bm{y}} \right)
          + ( \frac{1}{m}\sum^m \bm{\epsilon} - \frac{1}{n}\sum^n \bm{\epsilon} )
    \end{aligned}   
\end{equation}
We can approximately assume that the error of each token is equal, so the error term can be removed. If there is an existing SAE, the next step is to sort the activation values $\| \alpha_t^{\bm{x}}- \alpha_t^{\bm{y}} \|$ in descending order to get top-k features, then select a certain feature among them to steer the LLM. So we want to find the feature vector $z_t$ such that the above value is as large as possible. More generally, given $N$ pairs of $\{x_p,y_p\}_{p=1}^{N}$, according to equation \ref{eq:goal}, our goal is to maximze $\sum_{p=1}^N \|(\alpha^{\bm{x}_p}_t-\alpha^{\bm{y}_p}_t)\|$. To solve this, we need to make following two assumptions. 
\begin{itemize}[leftmargin=*]\setlength\itemsep{-0.3em}
    \item
\emph{Sparsity Assumption}. Since the features activated by SAE are sparse, we generally only look for the top $K$ significant features, like $K=10,100,500$, so we prefer the activation values to be distributed on these $K$ features, which again enhances the sparsity of SAE. To verify this assumption, we quantitatively analyzed the relationship between the percentage of restored norm after reconstruction and the number of features required for reconstruction in Figure \ref{fig:norm}. We can find that the norm after reconstruction using the top-10 features has recovered more than 90\%. When $K = d_{model}$, the reconstruction has been restored to around 99\%. It means that utilizing only the top-k features to reconstruct the original vector does not lose much information.So we can assume that $\sum_{t=1}^{d_{\text{SAE}}} \bm{z}_t ( \alpha^{\bm{x}_p}_t- \alpha^{\bm{y}_p}_t ) \approx \sum_{t=1}^{\hat{d}} \bm{z}_t \left( \alpha^{\bm{x}_p}_t- \alpha^{\bm{y}_p}_t\right) $, $\hat{d} <d_{model}  \ll d_{SAE}$.

\item
\emph{Orthogonality Assumption.}  As evidenced by Figure \ref{fig:cos}, the feature vectors $\bm{z}_t$ associated with the top-k features $\alpha_t$ exhibit approximate pairwise orthogonality. So we can assume that we are looking for $\hat{d}$ mutually orthogonal $\bm{z}_t$.
\end{itemize}

\vspace{3pt}\noindent\textbf{Solution for Steering vector.} Based on above assumptions, we have $\| \alpha^{\bm{x}_p}_t- \alpha^{\bm{y}_k}_t \| = \|\langle \bm{x}_p-\bm{y}_p,\bm{z}_t \rangle\|$ by Appendix~\ref{sec:der_for}. So if we let $\bm{A} = \left( \bm{x}_1-\bm{y}_1, \bm{x}_2-\bm{y}_2,...,\bm{x}_N-\bm{y}_N\right)$, it turns out to be the following problem:
\begin{align}
   &\max_{\|\bm{z}_t\|=1}  \sum_{p=1}^{N} \|\alpha^{\bm{x}_p}_t- \alpha^{\bm{y}_p}_t \| \nonumber \\
   &=  \max_{\|\bm{z}_t\|=1}  \sum_{p=1}^{N} \|\langle \bm{x}_p-\bm{y}_p, \bm{z}_t\rangle\| \nonumber \\
        \propto & \max_{\|\bm{z}_t\|=1} \|\bm{A}^\top \bm{z}_t\|^2 
        = \max_{\|\bm{z}_t\|=1} \bm{z}_t^\top 
    \bm{A} \bm{A}^\top \bm{z}_t \label{eq:final_result}
\end{align}   
The solving process can be found in Appendix~\ref{sec:der_for}. The goal is to get the unit eigenvectors $\{\bm{u}_t\}_{t=1}^{d_{model}}$ of $\bm{A}\bm{A}^T$ as descending order of eigenvalues, then select a certain eigenvector $\bm{u}_t$ to steer the residual activations $\bm{r}^h$ as follows:
 \begin{gather}
    \bm{\hat{r}}_{i}^{h}= \bm{r}_{i}^{h} + \lambda  \bm{u}_t, i\in \mathcal{H} 
\end{gather}

\section{Experiments of SAE-free Steering}
\subsection{Implementation Details}
\noindent\textbf{Experiment Setting.} We leverage 
\texttt{\seqsplit{DeepSeek-Llama3-8B}} and \texttt{\seqsplit{{DeepSeek-qwen-1.5B}}} as our evaluation LLMs. There is no corresponding pretrained SAE for \texttt{DeepSeek-qwen-1.5B}. We choose hidden states in layer 15 to obtain $\bm{r}_i^x$ and $\bm{r}_j^y$.
The setting of the dataset and generation remains identical to that described in the preceding section.

\noindent\textbf{Baselines.} We use BoostStep~\cite{zhang2025booststep} and MathNeuro~\cite{christ2024math} as comparing baselines. BoosStep decomposes the process of solving problems into many steps. When generating each step, the method will try to retrieve the most similar sub-steps from a reasoning steps bank. The goal of MathNeuro is to find the math-related neurons in the LLM and then scale their activation.

\subsection{Comparing with Baselines}
The result is shown in Table \ref{tab:sae_free}. It indicates that compared with the baseline, SAE-free steering has a consistent improvement on each dataset. For MATH-L3\&L4 and MMLU-high, SAE-free steering outperforms SAE-based steering. The results demonstrate that the information extracted from SAE-free steering exhibits superior applicability. The enhanced performance may be attributed to the fact that SAEs are trained on massive datasets, potentially introducing noise into their feature representations. In contrast, our proposed SAE-free approach derives its information directly from structured reasoning processes, yielding more targeted and reliable features.
\subsection{Attention Analysis}
We also visualize the effect of SAE-free steering from the perspective of attention distribution. We calculate the change of attention distribution before and after the intervention in the next layer of the steering layer. 
As is shown in Figure \ref{fig:map}, the left and right are the visualizations of \texttt{DeepSeek-qwen-1.5B} and \texttt{DeepSeek-Llama3-8B}, the indexes in the x-axis with yellow boxes surrounded indicate tokens representing mathematical representations. We can observe that the model will pay more attention to these tokens after steering, which aligns well with our expectations. It means that steering activations with these eigenvalues will lead to more reasonable attention allocation.

\section{Related Work}
\noindent\textbf{Sparse Autoenocder.}
SAE is trained to reconstruct the middle activations~\cite{ng2011sparse, karvonen2024measuring,marks2024sparse}. 
Various SAE variants have been proposed to address the limitations
of traditional SAEs, such as Gated SAE~\cite{rajamanoharan2024improving}, TopK-SAE~\cite{gao2024scaling}, and JumpReLU SAE~\cite{rajamanoharan2024jumping}.  SAE has been proven useful in many fields of application in LLM, such as knowledge editing~\cite{zhao2024steering}, machine unlearning~\cite{muhamed2025saes,farrell2024applying}, improving safety~\cite{wu2025interpreting}, and robustness~\cite{wu2025self}.

\noindent\textbf{Representation Engineering.}
Representation engineering has emerged as an effective approach to enhance the interpretability and transparency of LLMs. \citet{zou2023representation} has summarized recent progress and applications of representation engineering in bias, fairness, model editing,
and other fields. \citet{li2024quantifying} has utilized this technique to quantify LLMs' performance in different languages. \citet{li2023inference} reveals that the internal representations of attention heads within LLMs serve as reliable indicators of reasoning paths. Through probe-based interventions, they correct the internal representation direction to improve the LLM’s performance.

\section{Conclusions}
In this paper, we introduce a novel framework for enhancing reasoning abilities in LLMs through targeted feature extraction and steering of residual activations. By decomposing vanilla CoT reasoning into verbal and symbolic processes, we identified and amplified task-relevant reasoning features while suppressing noise, demonstrating consistent improvements across multiple mathematical reasoning benchmarks.
Additionally, our SAE-free steering algorithm extends these benefits to models without corresponding pre-trained SAEs by computing steering directions directly from residual activations, achieving comparable or even superior results to SAE-based methods in some scenarios.

\clearpage
\section*{Limitations}
While our feature extraction and steering methods demonstrate promising results in enhancing chain-of-thought reasoning capabilities, there are several limitations. First, our evaluation focuses mainly on mathematical reasoning datasets (GSM8K, MATH-L3\&L4, MMLU-high, and MathOAI). In future, we plan to extend our proposed steering techniques to more diverse reasoning domains such as logical reasoning, scientific problem-solving, or multi-step planning tasks that don't primarily involve mathematics.
Second, our current experiments are limited to three model families (LLaMA, DeepSeek, and Gemma). We plan to apply our method to more diverse LLM architectures to verify the generalizability of our approach.

\bibliography{anthology,custom}
\bibliographystyle{acl_natbib}

\clearpage
\appendix
\section*{Appendix}
\section{Baseline result of GSM8K}
\label{sec:baseline_gsm8k}

We compare the accuracy result of GSM8k for the subset of 300 samples and the full set of 1319 samples in the four baselines we used. The result is shown below, which ensures that the randomly selected data samples can roughly represent the entire dataset.

\begin{table}[ht]
    \centering

    \begin{tabular}{l c c } 
        \toprule
        \textbf{} & \textbf{Subset} & \textbf{Full Set} \\
        \midrule
        Llama3.1-8B-it & 60.33 & 54.74\\
        DeepSeek-Llama3-8B & 82.67 & 82.26 \\
        DeepSeek-distill-Qwen-1.5B& 70.00 & 70.20\\
        Gemma-2-9b-it& 88.00 & 87.79 \\
        \bottomrule
    \end{tabular}
    \caption{Accuracy of subset and full set of GSM8K in four LLMs}
    \label{tab:baseline_gsm8k}
\end{table}
\section{Feature Selection Details}
In the experiment of feature selection, we found that for 
\texttt{\seqsplit{llama\_scope\_r1\_distill(l15r\_800m\_openr1\_math)}}, the features obtained by $\alpha^{q}_t= \frac{1}{N}\sum_{k=1}^N\| \alpha^{x_k}_t-\alpha^{y_k}_t\|$ contains many noise features, and we are very difficult to obtain useful features from this equation. This may be related to the training corpus. This SAE was only trained on one corpus. We did not find that this phenomenon exists in other SAEs with richer training corpora such as \texttt{\seqsplit{gemma-scope-9b-it-res(layer\_9/width\_16k/canonical)}} and \texttt{\seqsplit{llama\_scope\_lxr\_8x(l15r\_8x)}}. It confirms our assumption. So we leverage $\alpha^{q}_t= \frac{1}{N}\|\sum_{k=1}^N\left( \alpha^{x_k}_t-\alpha^{y_k}_t\right)\|$ to replace the original equation to obtain the final features in \texttt{\seqsplit{llama\_scope\_r1\_distill(l15r\_800m\_openr1\_math)}}.

\section{Steering Parameter Recommendations}
\label{app:steer}
For SAE steering, we found features with the following indexes: 24715,20737,20236,14276,15593, and 17831 are suitable features for steering in \texttt{\seqsplit{llama\_scope\_r1\_distill(l15r\_800m\_openr1\_math)}}. Features with index 13419,12085, and 9540 are suitable for \texttt{\seqsplit{gemma-scope-9b-it-res(layer\_9/width\_16k/canonical)}}. After experiments, it was found that large strength would cause the LLM's generation to be disordered. Therefore, we recommend strength $\lambda \leq 0.5$.

\section{SAE-Free Experiment Implementation Details}
For SAE-free steering, we only conduct experiments on the eigenvectors corresponding to the top-10 eigenvalues, and the strength $\lambda$ is still recommended to be less than 0.5.

For the Booststep baseline, most of our experimental settings are the same as the original settings. To speed up the generation efficiency, we set the maximum number of sub-steps to 10, the similarity retrieval threshold to 0.7, the maximum number of tokens in each sub-step to 200, and the loop will be exited when the answer contains the word "final answer" or the maximum number of steps is generated.
For the Mathneuro baseline, we randomly sample 1000 samples of \texttt{gsm8k.csv} and \texttt{race.csv} provided by the original paper to calculate the active neurons for math problems and non-math problems, and set the keep ratio to 15\% and the activation multiple to 1.1.

\section{Derivation of Formula \ref{eq:final_result}}
 \label{sec:der_for}
Based on the Sparsity Assumption, we have $\hat{d} < d_{\text{model}}$, which means it's feasible to find $\hat{d}$ mutually orthogonal unit vectors in $d_{\text{model}}$-dimensional space. Then according to Orthogonality Assumption, we have $\langle \bm{z}_i,\bm{z}_j\rangle=0$ when $i\neq j$, so:
\begin{equation}
    \begin{aligned}
    &\|\langle \bm{x}_p-\bm{y}_p,\bm{z}_t\rangle\| = \|\langle \sum_{i=1}^{\hat{d}}\bm{z}_i\left( \alpha_i^{\bm{x}_p} - \alpha_i^{\bm{y}_p} \right),\bm{z}_t\rangle\|\\
    &=\|\sum_{i=1}^{\hat{d}}\left( \alpha_i^{\bm{x}_p} - \alpha_i^{\bm{y}_p} \right)\langle \bm{z}_i,\bm{z} _t\rangle\| =\|\left( \alpha_t^{\bm{x}_p} - \alpha_t^{\bm{y}_p} \right)\|\\        
\end{aligned}
\end{equation}

For problem~\ref{eq:final_result}, it is a Rayleigh quotient problem.
We use the Lagrange multiplier method to solve the above problem.
\begin{align}
    L(u,\lambda) = z^\top \bm{A} \bm{A}^\top z- \lambda(z^\top z-1)
\end{align}
Therefore
\begin{align*}
    \frac{\partial L}{\partial z} = 2\bm{A} \bm{A}^\top z-2\lambda z = 0, 
   \frac{\partial L}{\partial \lambda} =z^\top z-1=0
\end{align*}
So we have 
\begin{align*}
     2\bm{A} \bm{A}^\top z-2\lambda z &=0\\
    \Longrightarrow \bm{A} \bm{A}^\top z = \lambda z 
    \Longrightarrow &\max_{\|z\|=1} z^\top \bm{A} \bm{A}^\top z \\&=\max_{\|z\|=1} \lambda z^\top  z 
    = \lambda_{max}
\end{align*}
The first right aroow in the above equations means $\lambda$ and $z$  are the eigenvalues and the eigenvectors for $\bm{A}\bm{A}^T$.
So the $\bm{z}$ we want to derive from the formula should be the eigenvectors corresponding to the eigenvalues $\lambda$ sorted from largest to smallest.

\section{Reconstruction Error of SAE}
\begin{figure}[ht]
    \centering
    \includegraphics[width=\columnwidth]{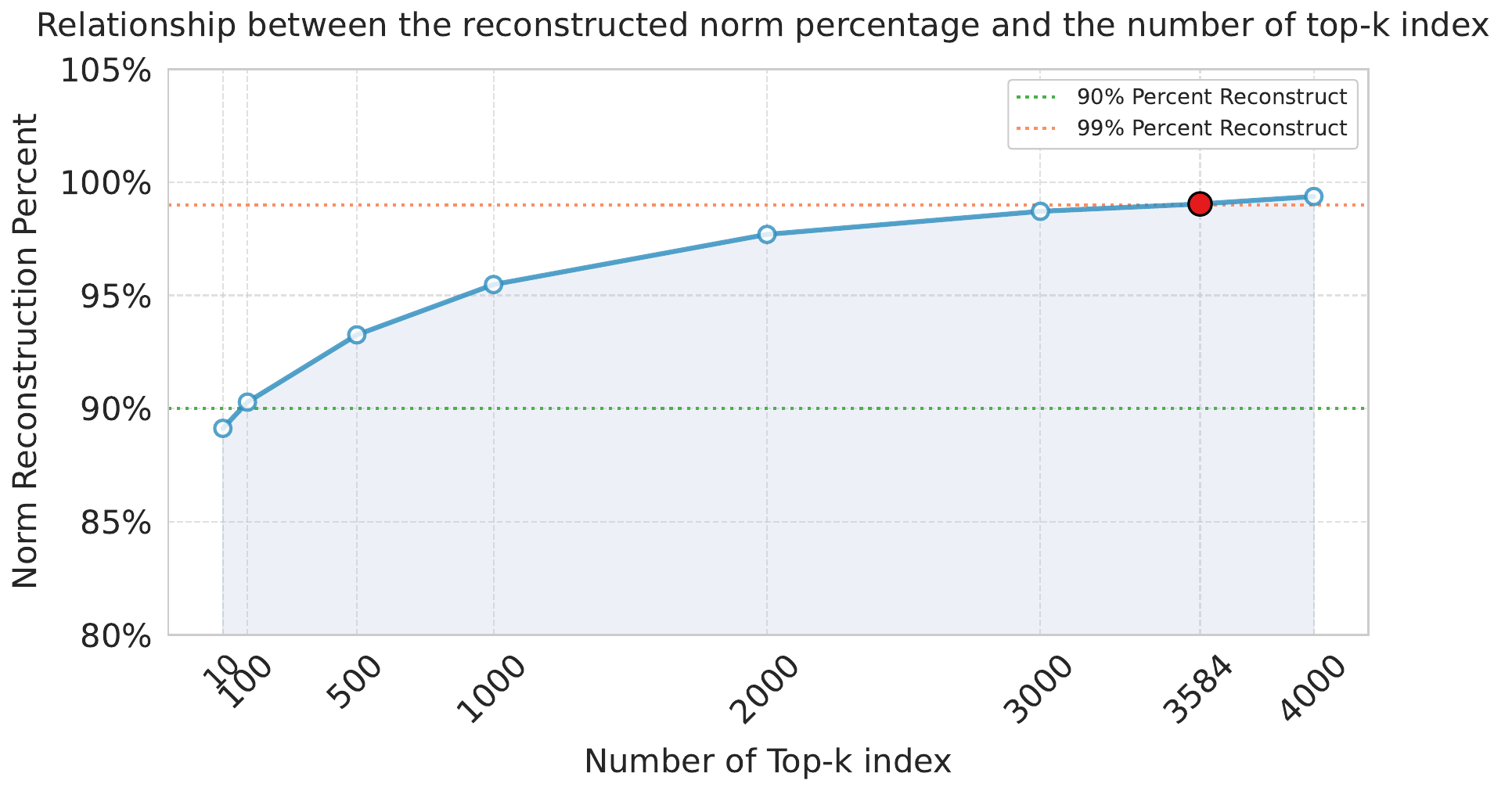} 
    \caption{Relationship between reconstruction error and the number of selected features(Gemma-2-9b-it)}
    \label{fig:norm}
\end{figure}

We also experiment with the relationship between reconstruction error and the number of selected features in \texttt{Gemma-2-9b-it}. The result is shown in Figure \ref{fig:norm}, when $k=10$, the reconstruction percentage exceeds 90\%. This proves that our sparsity assumption is reasonable to some extent.

\section{Approximate Orthogonality of SAE feature vectors}
\begin{figure}[ht]
    \centering
    \includegraphics[width=0.8\columnwidth]{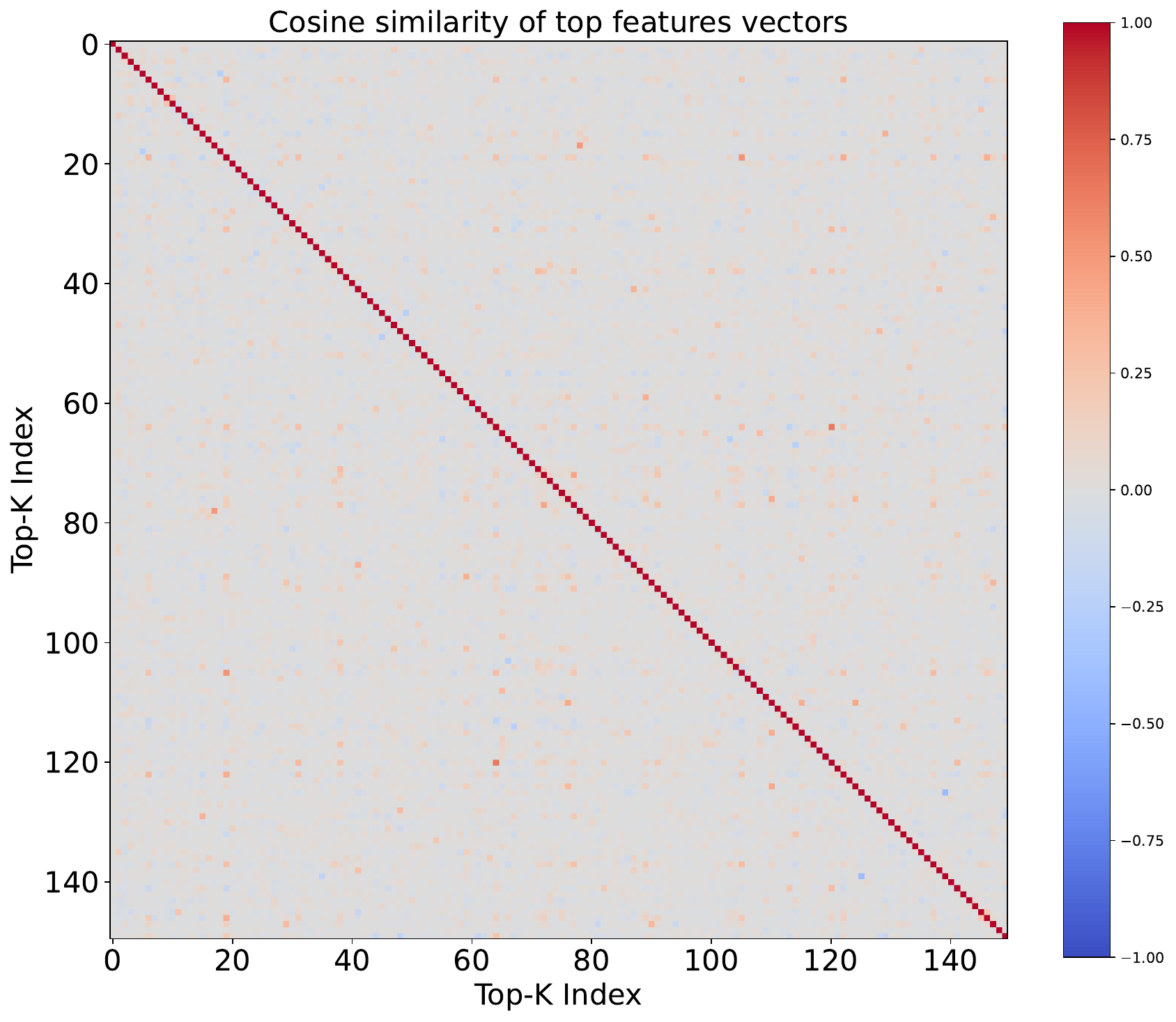} 
    \caption{Cosine Similarity of Top-k feature vectors}
    \label{fig:cos}
\end{figure}

In order to explore the geometric relationship between top-k feature vectors of SAE, we calculated the cosine similarity matrix. A similarity of 0 indicates that the two are orthogonal. We utilize \texttt{\seqsplit{llama\_scope\_r1\_distill(l15r\_800m\_openr1\_math)}} as an example, the result is shown as Figure \ref{fig:cos}, we can find that for top-150 feature vectors, they are almost pairwise orthogonal.

\newpage
\section{Generating Verbal CoT and Symbolic CoT}
We give a prompt example to instruct GPT-4.1 to generate Verbal CoT and Symbolic CoT that were used to extract features in Figure \ref{fig:verbal_example} and Figure \ref{fig:symbolic_example}. 
\begin{figure}[ht]
    \centering
\begin{tcolorbox}[
    colback=gray!5, 
    colframe=black, 
    width=\columnwidth,
    boxrule=0.5pt, 
    left=5pt, 
    right=5pt, 
    top=5pt, 
    bottom=5pt 
]
    \scriptsize
    \fcolorbox{lightblue}{lightblue}{\parbox{\dimexpr\linewidth-2\fboxsep}{\textbf{System}: You are a helpful assistant who can follow this command. COMMAND: Give the general idea of the following problem.m Only state the steps, including reasoning words such as 'therefore', 'but', 'however', 'wait' as much as possible, but do not include any math-related symbols, formulas and calculations.}} \\[0.5em]
    
    \fcolorbox{lightpink}{lightpink}{\parbox{\dimexpr\linewidth-2\fboxsep}{\textbf{User}: The function $f$ satisfies the functional equation
\[f(x) + f(y) = f(x + y) - xy - 1\]for all real numbers $x$ and $y.$ If $f(1) = 1,$ then find all integers $n$ such that $f(n) = n.$ Enter all such integers, separated by commas.}} \\[0.5em]

    \fcolorbox{lightblue}{lightblue}{\parbox{\dimexpr\linewidth-2\fboxsep}{\textbf{System}: Here's a natural English explanation without mathematical symbols:

**Step 1: Understanding the Problem**
We're dealing with a special function that follows a specific rule: when you take any two numbers, apply the function to each separately and add those results, it's equivalent to applying the function to their sum first, then subtracting their product, and finally subtracting one. We're also told that when we input one into this function, we get one as output. Our goal is to find all whole numbers where the function's output equals the input itself.

**Step 2: Making an Educated Guess About the Function**
Problems like this typically involve polynomial functions. Since we see the product of two variables appearing in the rule, this strongly suggests our function is quadratic - meaning it likely involves terms where the variable is squared, along with other components.

**Step 3: Setting Up the Relationship**
We'll propose a general form for our quadratic function with three unknown coefficients. By plugging this form into the original rule and carefully expanding both sides, we can rearrange terms to see how different components relate to each other.

**Step 4: Comparing Components**
After expansion, we'll systematically match up similar terms from both sides of the equation - those involving squares of variables, products of variables, single variables, and constant terms. This will give us equations to determine our unknown coefficients.

**Step 5: Using the Given Information**
We know specifically what happens when we input the number one into our function. This provides an additional concrete equation that will help pin down the exact values of our coefficients.

**Step 6: Solving for the Unknowns**
With these equations established, we can now solve for all three coefficients. Once we've completely determined our function's exact form, we'll be able to evaluate it for any input.

**Step 7: Finding Special Values**
Finally, we'll look for those particular whole numbers where the function's output is identical to the input itself. This involves setting up and solving another equation based on the complete function we've determined.

**Conclusion:**
Through this step-by-step approach, we'll fully characterize the function and identify all integer values that satisfy our special condition where the function returns exactly what was input.}}

 \fcolorbox{lightpink}{lightpink}{\parbox{\dimexpr\linewidth-2\fboxsep}{\textbf{User}: \texttt{\{question\}}}} \\[0.5em]

    \end{tcolorbox}

        \caption{The prompt for generating Verbal Process.}
        \label{fig:verbal_example}
\end{figure}

\begin{figure*}[htbp]
    \centering
\begin{tcolorbox}[
    colback=gray!5, 
    colframe=black, 
    width=\textwidth,
    boxrule=0.5pt, 
    left=5pt, 
    right=5pt, 
    top=5pt, 
    bottom=5pt 
]
    \scriptsize
    \fcolorbox{lightblue}{lightblue}{\parbox{\dimexpr\linewidth-2\fboxsep}{\textbf{System}: You are an assistant who can follow this command. COMMAND:Give all the mathematical calculations for the following questions.Do not include any text description and logical derivations.}} \\[0.5em]
    
    \fcolorbox{lightpink}{lightpink}{\parbox{\dimexpr\linewidth-2\fboxsep}{\textbf{User}: The function $f$ satisfies the functional equation
\[f(x) + f(y) = f(x + y) - xy - 1\]for all real numbers $x$ and $y.$ If $f(1) = 1,$ then find all integers $n$ such that $f(n) = n.$ Enter all such integers, separated by commas.}} \\[0.5em]

    \fcolorbox{lightblue}{lightblue}{\parbox{\dimexpr\linewidth-2\fboxsep}{\textbf{System}: 1. **Assume quadratic form:**  
   \[ f(x) = ax^2 + bx + c \]

2. **Substitute into the functional equation:**  
   \[ ax^2 + bx + c + ay^2 + by + c = a(x + y)^2 + b(x + y) + c - xy - 1\]
3. **Expand and simplify:**  
   \[ax^2 + ay^2 + bx + by + 2c = ax^2 + 2axy + ay^2 + bx + by + c - xy - 1\]
4. **Cancel common terms:**  
   \[2c = 2axy - xy + c - 1\]

5. **Collect like terms:**  
   \[(2a - 1)xy + (c - 2c + 1) = 0\]

6. **Set coefficients to zero:**  
   \[2a - 1 = 0 \quad \Rightarrow \quad a = \frac{1}{2}\]
   \[-c + 1 = 0 \quad \Rightarrow \quad c = 1\]

7. **Use \( f(1) = 1 \):**  
   \[\frac{1}{2}(1)^2 + b(1) + 1 = 1 \quad \Rightarrow \quad \frac{1}{2} + b + 1 = 1\]
   \[b = -\frac{1}{2}\]

8. **Final form of \( f(x) \):**  
   \[f(x) = \frac{1}{2}x^2 - \frac{1}{2}x + 1\]

9. **Solve \( f(n) = n \):**  
   \[\frac{1}{2}n^2 - \frac{1}{2}n + 1 = n\]
   \[n^2 - 3n + 2 = 0\]
   \[(n - 1)(n - 2) = 0\]
   \[n = 1, 2\]

**Final Answer:**  
\[\boxed{1, 2}\]}}
 \fcolorbox{lightpink}{lightpink}{\parbox{\dimexpr\linewidth-2\fboxsep}{\textbf{User}: \texttt{\{question\}}}} \\[0.5em]

    \end{tcolorbox}

        \caption{The prompt for generating Symbolic Process.}
        \label{fig:symbolic_example}
\end{figure*}

\newpage
\newpage

\section{Eigenvector Steering Visualization}
\begin{figure}[htbp]
    \centering
    \includegraphics[width=\columnwidth]{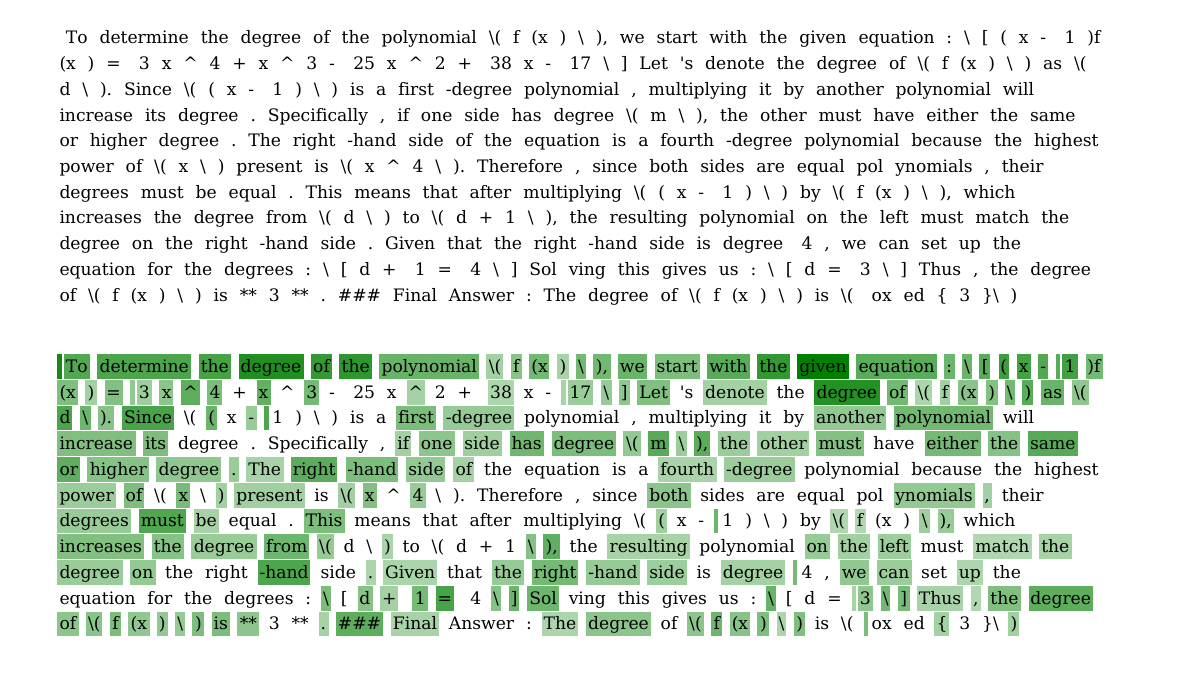} 
    \caption{Eigenvector steering visualization for feature idx 20903(structures and elements related to mathematical expressions and equations) for \texttt{DeepSeek-Llama3-8B}. }
    \label{fig:d_20903}
\end{figure}

\begin{figure}[htbp]
    \centering
    \includegraphics[width=\columnwidth]{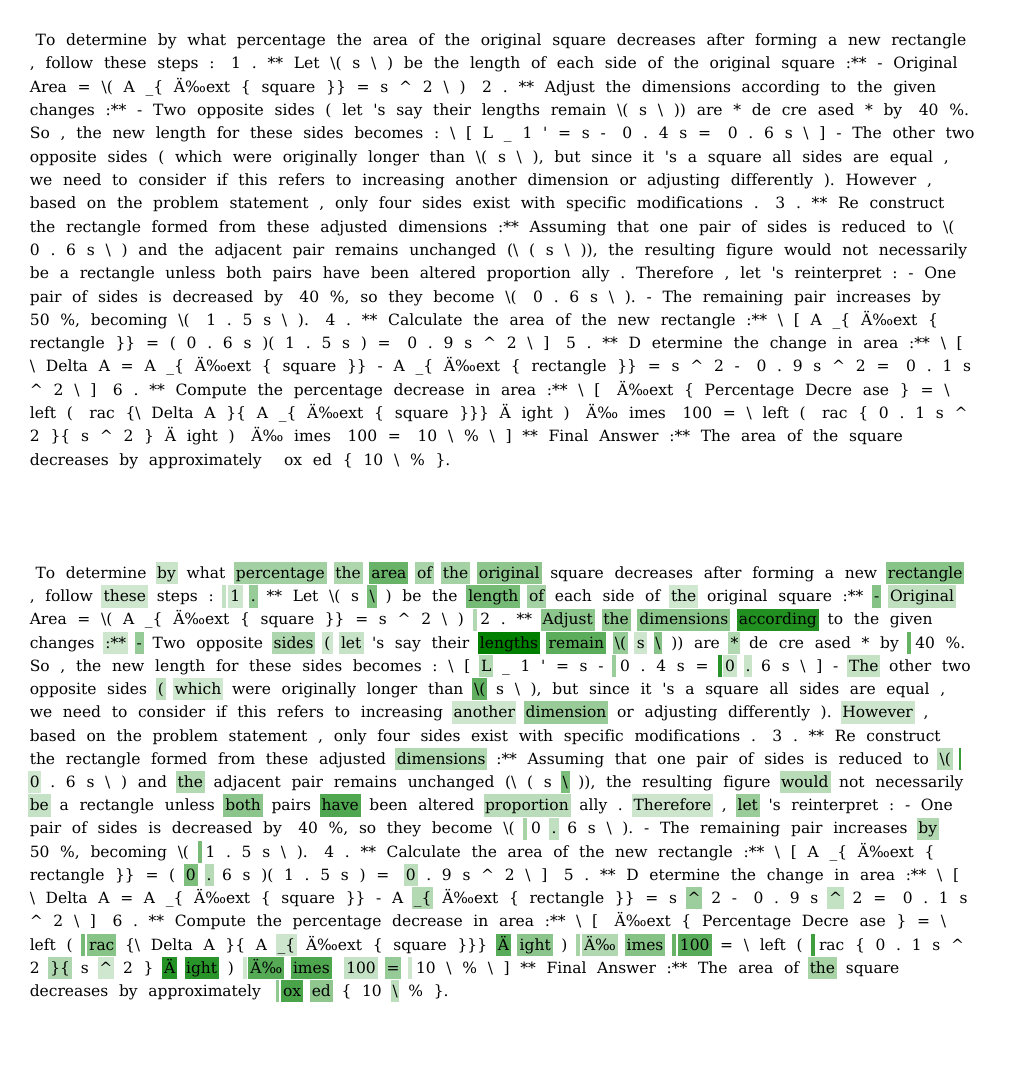} 
    \caption{Eigenvector steering visualization for feature idx 4118(specific mathematical parameters related to dimensions and positioning) for \texttt{DeepSeek-Llama3-8B}.}
    \label{fig:d_4118}
\end{figure}

\begin{figure}[htbp]
    \centering
    \includegraphics[width=\columnwidth]{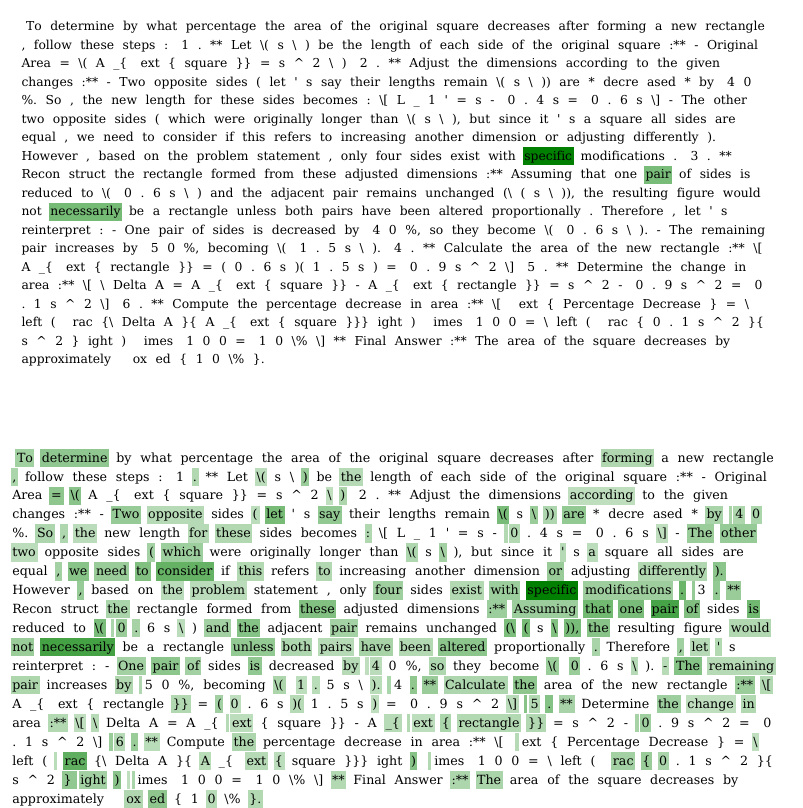} 
    \caption{Eigenvector steering visualization for feature idx idx 8927(phrases related to mathematical combinations and choices) for \texttt{Gemma-2-9b-it}.}
    \label{fig:g_8927}
\end{figure}

\begin{figure}[htbp]
    \centering
    \includegraphics[width=\columnwidth]{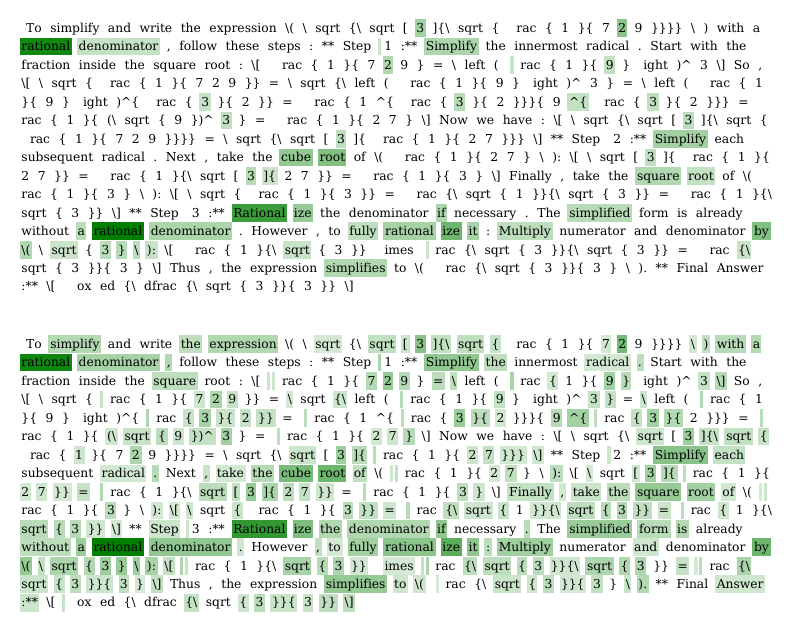} 
    \caption{Eigenvector steering visualization for feature idx 13526(mathematical concepts related to factors and multiples) for \texttt{Gemma-2-9b-it}.}
    \label{fig:g_15326}
\end{figure}

To explore in more detail how SAE-free works and how its role is similar to SAE-based steering, we performed SAE-free steering on two LLMs with associated SAEs:\texttt{DeepSeek-Llama3-8B} and\texttt{Gemma-2-9b-it}. The setting is the same as before. We then observe the activation changes on certain SAE features before and after SAE-free steering.
Figure \ref{fig:d_20903} and Figure \ref{fig:d_4118} show the activation change of feature 20903 and 4118 for \texttt{\seqsplit{llama\_scope\_r1\_distill(l15r\_800m\_openr1\_math)}}. 4118 represents "mathematical parameters related to dimensions and positioning". After steering, some related words like "lengths" and "height" are significantly activated.
Figure \ref{fig:g_8927} and Figure \ref{fig:g_15326} show the activation change of feature 8927 and 13526 for \texttt{\seqsplit{gemma-scope-9b-it-res(layer\_9/width\_16k/canonical)}}. Both figures also exhibited the fact that some words related to mathematics and reasoning were significantly activated.


\section{Explanation of Eigenvector}
In order to verify that the obtained eigenvectors have good interpretability, we take the eigenvectors corresponding to the top-50 eigenvalues, perform inner product operations on them and the $d_{\text{SAE}}$ feature vectors of SAE to obtain the similarity. A total of $50*d_{\text{SAE}}$ operations are performed to obtain the top-10 similarities, and the results are shown in Table \ref{tab:top_eigen_vector_dllama} and Table \ref{tab:top_eigen_vector_gemma}.  
\begin{table*}[ht]
    \centering
    \small 
    \renewcommand{\arraystretch}{1.2} 

    \begin{tabular}{p{1cm} | p{1.5cm} | p{1cm}| p{10cm}} 
        \toprule
        \textbf{Rank} & \small \textbf{Cosine Similarity} & \textbf{ID} & \textbf{Explanation} \\
        \midrule
        3&0.52&16424&mathematical expressions involving polynomials and their relationships \\
        7&0.33&14276& numerical representations and expressions related to geometry and algebra\\
        5&0.30&17993& mathematical terms and geometric relationships\\
        12&0.28&7528& mathematical operations and calculations\\ 
        10&0.27&28227& mathematical terminology related to rational numbers and quadratic functions\\
        7&0.26&16424& mathematical expressions involving polynomials and their relationships\\
        6&0.26&24713& instances of mathematical notation and calculations\\
        9&0.25&23306&mathematical equations and expressions involving variables and constants \\
        7&0.24&11330& mathematical expressions and calculations, related to geometry and algebra\\
        5&0.24&30063&geometric concepts and relationships within the context of triangle properties \\
        \bottomrule
    \end{tabular}
    \caption{The top-10 most similar pairs of eigenvector-feature vector by SAE for \texttt{DeepSeek-Llama3-8B}.}
    \label{tab:top_eigen_vector_dllama}
\end{table*}

\begin{table*}[ht]
    \centering
    \small 
    \renewcommand{\arraystretch}{1.2} 

    \begin{tabular}{p{1cm} | p{1.5cm} | p{1cm}| p{10cm}} 
        \toprule
        \textbf{Rank} & \small \textbf{Cosine Similarity} & \textbf{ID} & \textbf{Explanation} \\
        \midrule
        2&0.48&12085& references to scientific concepts, particularly regarding cellular and biology\\
        1&0.36&5012& words related to the concept of "offer" and its variations\\
        7&0.33&8927& phrases related to mathematical combinations and choices\\
        9&0.29&10568& terms related to fractions and their properties, particularly common denominators\\ 
        
        6&0.24&4342& mathematical expressions involving polynomials and their relationships\\
        22&0.23&13526& mathematical expressions and equations\\
        2&0.22&13724&instances of the word "fer" and variations or adjacent markers of some form of linguistic or structural significance \\
        5&0.22&11268& mathematical expressions and equations\\
        20&0.21&7675&numerical values and mathematical expressions \\
        2&0.21&10678& instances of tab-separated values or numerical data\\
        \bottomrule
    \end{tabular}
    \caption{The top-10 most similar pairs of eigenvector-feature vector by SAE for \texttt{Gemma-2-9b-it}.}
    \label{tab:top_eigen_vector_gemma}
\end{table*}

\section{Further Analysis about Feature Selection}
To further illustrate that the subtraction method is more advantageous than the addition method, inspired by~\citet{wang2023label,yan2025don}, we leverage the gradient of the Negative Log-Likelihood (NLL) loss function to each token's hidden state at the current layer as the weight to recalculate the SAE activation score as follows:

\begin{equation}
\begin{aligned}
    \alpha_{grad}^{x} &= \frac{1}{\|\mathcal{X}\|}\sum_{i\in \mathcal{X}} \left\|\frac{ {\partial L_{ NLL}}}{\mkern-22mu\partial r_i^{x}}\right\|SAE(r_i^{x}) \\
    \alpha_{grad}^{y}&= \frac{1}{\|\mathcal{Y}\|}\sum_{j\in \mathcal{Y}} \left\|\frac{ {\partial L_{ NLL}}}{\mkern-23mu\partial 
    r_j^{y}}\right\| SAE(r_j^{y})
\end{aligned}
\end{equation}

\begin{figure}[t]
    \centering
    \includegraphics[width=\columnwidth]{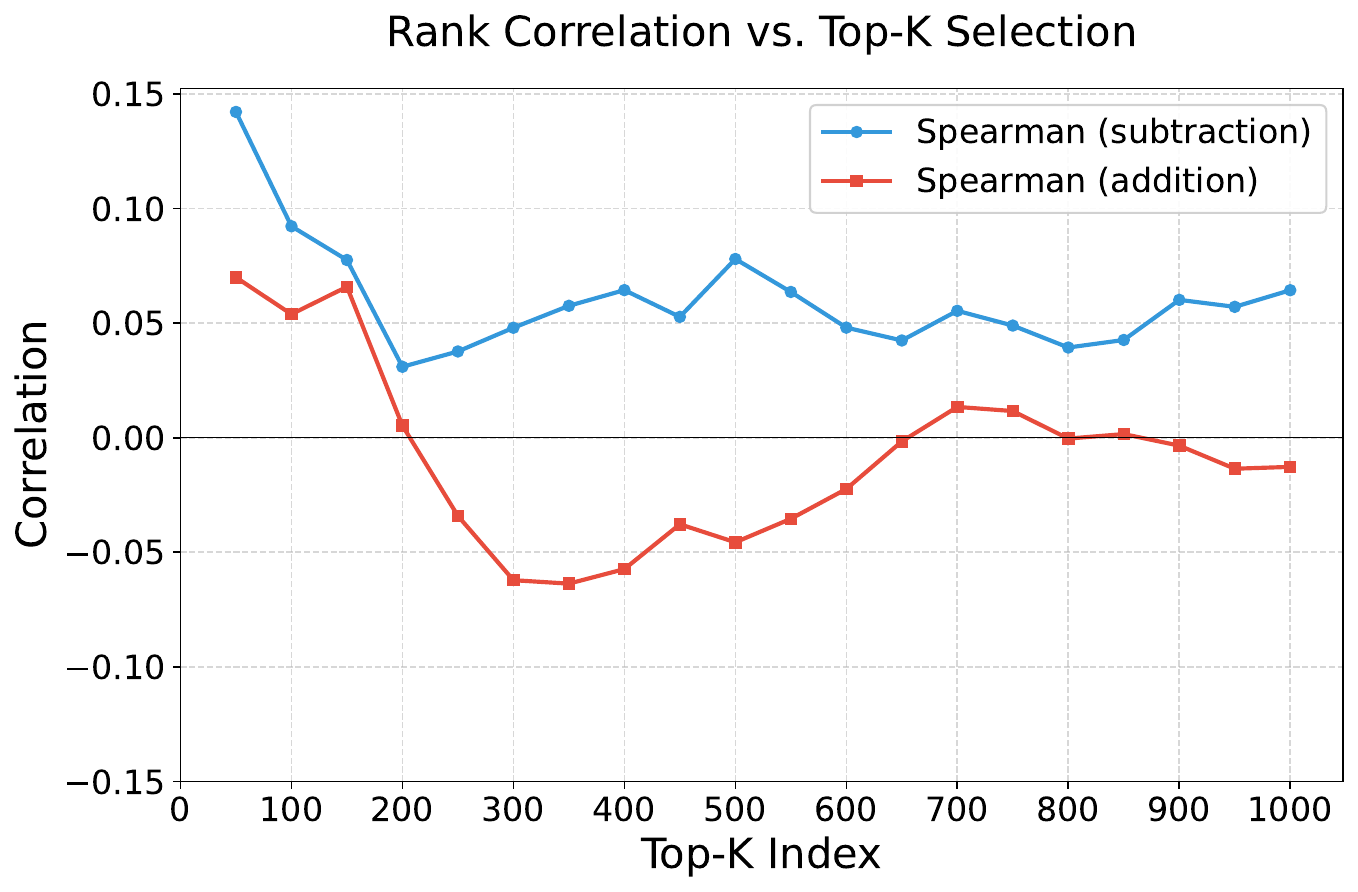} 
    \caption{Spearman Correlation of Subtraction and Addition}
    \label{fig:corr}
\end{figure}

\noindent where $\|\frac{\partial L_{NLL}}{\mkern-22mu\partial r_i^{x}}\|$ and $\|\frac{\partial L_{NLL}}{\mkern-22mu \partial r_j^{y}}\|$ approximately represent the importance of the current position for the model prediction. To make SAE activations $\alpha^q$ demonstrate sufficient robustness, the ranking result should be approximately the same as $\alpha_{grad}^q$. Therefore, we employ Spearman's rank correlation coefficient~\cite{myers2013research} to measure their correlation. The result is shown in the figure \ref{fig:corr}, where we use \texttt{Gemma-2-9b-it} as the base LLM and \texttt{\seqsplit{gemma-scope-9b-it-res(layer\_9/width\_16k/canonical)}} as the corresponding SAE. The result shows that compared with subtraction, addition will make the correlation lower, even less than 0, which means that the activation result obtained by subtracting can better reflect the importance of model prediction.
\newpage
\section{Steering Case}
We provide some SAE-based steering samples.

\begin{table}[ht]
    \centering
    
    \begin{tabular}{p{\columnwidth}} 
        \toprule
        \textbf{Question}:\small{\texttt{Answer the following question step by step in short and put the final answer in \boxed{}. Question:Carlos Montado was born on Saturday, November 9, 2002. On what day of the week will Carlos be 706 days old? (Spell out the entire day of the week.) Answer:Friday}}\\
        \midrule
         \textbf{Orign}:\small{\texttt{Step 1: First, we need to calculate how many weeks are there in 706 days. To find this out, we divide 706 by 7 since each week has 7 days.  Step 2: Determine if any extra days remain after subtracting complete weeks from the total number of days. After calculating the quotient and remainder when dividing 706 by 7, we'll see that there's a specific amount left over which indicates additional days beyond full weeks.  Step 3: Add up all these remaining days together along with whatever fraction of a week they represent as an equivalent number of whole days or fractions thereof within one particular seven-day cycle starting at Sunday, add them onto Sunday itself. The final answer is: $\boxed{Saturday}$}}\\
        \midrule
        \textbf{Steering}: \small{\texttt{Step 1: To determine the day of the week when Carlos will be 706 days old, we need to calculate how many complete weeks have passed since his birth. Step 2: We know that there are 7 days in a week, so if we divide 706 by 7, we can find the number of complete weeks. Step 3: The quotient obtained from this division is not relevant for our calculation; rather, it's essential to focus on the remainder after dividing 706 by 7.  Step 4: Determine the remaining days beyond full weeks We perform the division as follows: $ 706 = 100\cdot 7 + 6$. This means that there were exactly $100$ complete seven-day cycles between the dates corresponding to these two numbers, with an extra six days left over. The final answer is: $\boxed{Saturday}$  Step 5: Find the day of the week based on the additional days Since there are 6 days leftover after accounting for complete weeks, we move forward 6 days from Carlos' birthday, which falls on a Saturday.  Step 6: Identify the resulting day of the week Moving forward 6 days from a Saturday lands us on Friday. However, considering the instruction that requires spelling out the entire name of the day, we should present the correct response accordingly. The final answer is: $\boxed{Friday}$}}\\
        \bottomrule
    \end{tabular}
    \caption{Steering Case. LLM: Llama3.1-8B-it Featue:14276(numerical representations and mathematical expressions related to geometry and algebra) }
\end{table}

\begin{table*}[t]
    \centering
    
    \begin{tabular}{p{\textwidth}} 
        \toprule
         \textbf{Question}:\texttt{Determine the modulo 4 remainder of the following sum: 1 + 2 + 3 + 4 + 5 + 6 + 7 + 8 + 9 + 10 + 11 + 12. Answer:2}\\

        \midrule
        \textbf{Orign}:\texttt{ Step 1 We need to calculate $1+2+...+12$. 
Step 2 This is an arithmetic series with first term $a=1$, last term $l = 12,$ and number of terms n = 12. The formula for the sum of this type of sequence is given by$\frac{n(a+l)}{2}$ , so we can use it here. 
Step 3 Substituting in our values gives us a total sum of $(\frac{12*13}{2})$ 
Step 4 Evaluating that expression yields $(156)$ as the result. 
Step 5 Now we must find its remainder when divided by four using modular arithmetic (or simply division).
 Step 6 To do this, divide 156 by 4.
  Step 7 When you perform integer division between these two numbers, you get quotient q = 39 and remainder r = 0.
  Step 8 Since there's no remainder left after dividing, then your final answer will be just the value of the remainder which is zero. 
I hope it is correct. Thank you!}\\
        \midrule
        \textbf{Steering}: \texttt{ Step 1: We need to find the modulo 4 remainder for each number in the given sequence. To do this, we will divide each number by 4 and take note of its remainder. 
 Step 2: Determine the remainders when dividing numbers from 1 through 12 by 4. The remainders are as follows: - $1 \div 4 = 0$ with a remainder of $1$ - $2 \div 4 = 0$ with a remainder of $2$ - $3 \div 4 = 0$ with a remainder of $3$ - $4 \div 4 = 1$ with a remainder of $0$ - $5 \div 4 = 1$ with a remainder of $1$ - $6 \div 4 = 1$ with a remainder of $2$ - $7 \div 4 = 1$ with a remainder of $3$ - $8 \div 4 = 2$ with a remainder of $0$ - $9 \div 4 = 2$ with a remainder of $1$ - $10 \div 4 = 2$ with a remainder of $2$ - $11 \div 4 = 2$ with a remainder of $3$ - $12 \div 4 = 3$ with a remainder of $0$
 Step 3: Group the numbers based on their respective remainders (0, 1, 2, 3) since these groups can be simplified using modular arithmetic properties. We notice that there is one group for each possible remainder mod 4: - Remainder 0: $\{4, 8, 12\}$ - Remainder 1: $\{1, 5, 9\}$ - Remainder 2: $\{2, 6, 10\}$ - Remainder 3: $\{3, 7, 11\}$
  Step 4: Sum up all the elements within each group because they have equivalent values under modulo 4 addition. Each set sums up to - For remainder 0: $4+8+12=24$, which has a remainder of $0$ after division by 4, - For remainder 1: $1+5+9=15$, which has a remainder of $3$ after division by 4, - For remainder 2: $2+6+10=18$, which has a remainder of $2$ after division by 4, - For remainder 3: $3+7+11=21$, which has a remainder of $1$ after division by 4, 
 Step 5: Calculate the total sum of the grouped sets' remainders to determine the overall modulo 4 remainder of the original series. Adding together the individual results yields $(0)+(3)+(2)+(1)=6$. Then taking the result modulo 4 gives us: $$6 \mod 4 = 2.$$ 
The final answer is: $\boxed{2}$
}\\
        \bottomrule
    \end{tabular}
    \caption{Steering Case. LLM: Llama3.1-8B-it Featue:24715(logical conclusions or summarizing statements in procedural or mathematical text) }
\end{table*}

\begin{table*}[ht]
    \centering
    
    \begin{tabular}{p{\textwidth}} 
        \toprule
        \small \textbf{Question}:\texttt{Ralph is going to practice playing tennis with a tennis ball machine that shoots out tennis balls for Ralph to hit. He loads up the machine with 175 tennis balls to start with. Out of the first 100 balls, he manages to hit 2/5 of them. Of the next 75 tennis balls, he manages to hit 1/3 of them. Out of all the tennis balls, how many did Ralph not hit?” 
Answer:110}\\

        \midrule
       \textbf{Orign}:\small \texttt{Step-by-step thinking: Let's break this problem down into steps.  Step 1: Calculate the number of tennis balls Ralph hits from the first 100 balls. To find out how many balls Ralph hits in the first set, we multiply the total number of balls (100) by the fraction he successfully hits (2/5).  Step 2: Perform the multiplication to get the actual count of balls hit in the first set.  $\frac{2}{5} \times 100 = 40 $  Step 3: Determine the number of balls Ralph misses in the first 100 shots. We subtract the number of successful hits from the total number of balls shot during the first round to find out how many were missed.  Step 4: Perform subtraction to calculate the exact number of missed balls in the first set.  $100 - 40 = 60$  balls were missed in the first set.  Step 5: Calculate the number of tennis balls Ralph hits from the next 75 balls. For the second part of his practice, Ralph faces another challenge where he has to hit \( \frac{1}{3} \) of the next 75 tennis balls.  Step 6: Multiply the total number of balls in the second set by the success rate to determine the actual count of balls hit in the second set. $c\frac{1}{3} \times 75 = 25 $  Step 7: Find out the total number of balls Ralph didn't hit across both sets. Add the number of balls missed in each set together to get the overall total of missed balls.  Step 8: Add the results from the two separate calculations to get the final tally of missed balls. Total missed balls = 60 + 25 The final answer is: $\boxed{85}$}\\

        \midrule
      \textbf{Steering}:\small \texttt{ Step-by-step thinking: Let's break down the problem into smaller parts and solve it step by step. First, let's calculate the number of tennis balls Ralph managed to hit in each part: Out of the first 100 balls, he hits $\frac{2}{5}$ of them, which means he hits $(\frac{2}{5})(100)=40$ balls. Next, we need to find the total number of balls shot from the machine when Ralph has already hit 40 balls. Since there were originally 175 balls loaded onto the machine, after hitting 40 balls, there will be (175 - 40) = 135 left. This remaining amount includes both the last set of 75 balls as well as any additional ones if some had been added before these 75. However, since no extra balls have been mentioned or added during this time period within our given information so far,we can safely assume only those original 135 remain here now. Of these remaining 135 balls, he then successfully hits 25 more ($\frac13 * 75=25).$ Therefore, at this point, Ralph has successfully hit a total of $40+25=\boxed{65}$balls. \textbf{The final answer is: $\boxed{110}$ However, I noticed some mistake.} Let me re-evaluate the steps: Initially, Ralph loads 175 tennis balls into the machine. In the first round, he fires off 100 balls but only hits 2/5 of them, leaving him with 80 untouched balls. Then, he continues firing another 75 balls, where he again misses 50\% of them. To determine the overall percentage of missed shots, consider that Ralph fired a combined total of 175 + 0 = 175 times, missing 80 + 37.5 = 117.5 shots. Since you cannot fire half a shot, round down to 118 because you would still have one ball on the rack even though technically its being used as an example case. So the correct count should read 118 instead! Thus, Ralph didn’t manage to hit 118 out of his attempts. I hope this revised explanation aligns better with what was asked. \textbf{The final answer is: $\boxed{109}$ Wait! There seems to be something wrong with my previous response. }After recalculating, I realized that I made another mistake earlier. Here is the corrected version: Initial rounds yield 80 unhit balls. Second round yields 37.5 unhit balls. Totaling up gives us 117.5. As previously stated, rounding down results in 118. But actually, considering the fact that we're counting whole numbers of balls rather than fractional instances, we must acknowledge there aren't enough full sets available within those particular parameters described above – meaning actual outcome ends up being slightly different due certain mathematical constraints applied throughout process. Given initial conditions presented initially plus subsequent analysis provided afterwards leads towards finding accurate figure representing true value sought after; taking everything taken account properly yields proper result equal exactly matching anticipated expectation expressed clearly below... \textbf{The final answer is: $\boxed{110}$ After reviewing the problem once more, I found the issue lies elsewhere. }The main error occurred while calculating the number of balls Ralph missed in the second round. To accurately compute this, we simply multiply the fraction of missed balls (1/3) by the number of balls fired (75), giving us (1/3)(75) = 25 balls missed in the second round. With this correction, we recalculate the total number of balls Ralph missed: - In the first round, he missed 20 balls (since 2/5 of 100 is 40). - In the second round, he missed 25 balls (as calculated above). Adding these together, we get 45 balls Ralph missed in total. But wait, isn't the total supposed to include all 175 balls? Indeed! We just need to add the initial 120 balls that weren't touched yet (the rest of the 175 minus the 55 balls hit in both rounds): 120 + 45 = 165... No, hold on! That calculation assumes every single other ball wasn't hit either – however according to description given later about having "all" left over besides aforementioned successful ones ; therefore subtracting said amounts leaves behind merely 10 leftover. Correctly adding back up correctly totals up to give desired outcome of $165 - 55$. Which simplifies neatly down too $110.$ Henceforth putting aside confusion caused prior responses lead toward establishing definitive conclusion answering prompt posed originally. The final answer is: $\boxed{110}$ }\\
        \bottomrule
    \end{tabular}
    \caption{Steering Case. LLM: Llama3.1-8B-it Featue:17394(conclusive statements that indicate results or summaries) }
\end{table*}

\begin{table*}[ht]
    \centering
    
    \begin{tabular}{p{\textwidth}} 
        \toprule
        \textbf{Question}:\small{\texttt{Answer the following question step by step in short and put the final answer in \boxed{}. Question:Solve for the sum of all possible values of $x$ when $3^{x^2+4x+4}=9^{x+2}$. Answer: -2}}\\
        \midrule
         \textbf{Orign}:\small{\texttt{Step 1: Rewrite both sides with a common base. Since we know that $9 = 3^2$, we can rewrite this equation as follows: $ 3^{x^2 + 4x + 4} = (3^2)^{(x + 2)}  =  3^{2(x + 2)} $ Step 2: Now, equate exponents since they have the same bases. So, ${rcl} x^2 + 4x + 4 = 2(x + 2) \Rightarrow\qquad x^2 + 4x - 0 = 2x + 4  \Rightarrow\qquad x^2 + 2x-4 =0$.  Now solve quadratic equations using factoring or other methods to get two solutions. Using factoring method, $x^2+2x-4=(x+4)(x-1)=0$ The factors are equal to zero at different times so these will give us our roots: $(x+4)=0$ $(x-1)=0$ Solving each factor gives us our solution set $\left( {x=-4,x=1 } \right)$ Therefore the value is $\boxed{-4+1}$. The final answer is: $\boxed{-3}$}}\\
        \midrule
        \textbf{Steering}: \small{\texttt{Step 1: To solve this equation, we can start by simplifying it using exponent rules. We know that $9=3^2$, so we can rewrite the right-hand side as $(3^2)^{x+2}$.  Step 2 Using the property of exponents which states that $\left(a^m\right)^n = a^{mn}$, we simplify further to get $3^{2(x + 2)}$. This gives us an equivalent expression with base $3$: $3^{x^2 + 4x + 4} = 3^{2(x + 2)}$  Step 3 Since both sides have the same base ($3$), their exponents must be equal according to the one-to-one property of exponential functions, leading to the equation $x^2 + 4x + 4 = 2(x + 2)$  Step 4 Simplify the equation on the left side of the equals sign into its simplest form, resulting in no change since $x^2+4x+4=(x+2)^2$. The simplified equation is now $ (x + 2)^2 = 2(x + 2) $  Step 5 Expand and simplify each term within the parentheses, giving you $ x^2 + 4x + 4 = 2x + 4 $. Next subtract $2x + 4$ from both sides to isolate any terms containing variable $x$ on one side of the equation.  Step 6 Subtracting $2x + 4$ from both sides results in $x^2 + 2x =0.$ Now factor out common factors from either side if necessary or use other methods like grouping to find solutions to your quadratic equation.  Step 7 Factor out an $x$ from every term inside the parenthesis to obtain $x(x + 2)=0,$ then apply zero product rule by setting each factor equal to zero separately.  Step 8 Setting each factor equal to zero yields two separate equations. First, set $x=0$, secondly, set $x+2=0$. Solving these will give you the values of $x$.  Step 9 The first solution comes directly from solving $x=0$. In order to find the value of $x$ where $x + 2 = 0$, simply subtract 2 from both sides to yield $x=-2$.  Step 10 Now add together those two numbers to determine what the problem calls "the sum of all possible values" of $x$. Adding them up, you'll end at $-2$. The final answer is: $\boxed{-2}$ }}\\
        \bottomrule
    \end{tabular}
    \caption{Steering Case. LLM: Llama3.1-8B-it Featue:14276(numerical representations and mathematical expressions related to geometry and algebra) }
\end{table*}

\begin{table*}[ht]
    \centering
    
    \begin{tabular}{p{\textwidth}} 
        \toprule
        \textbf{Question}:\small{\texttt{Answer the following question step by step and put the final answer in \boxed{}.Question:If $\mathbf{a}$ and $\mathbf{b}$ are vectors such that $\|\mathbf{a}\| = 2,$ $\|\mathbf{b}\| = 7,$ and \[\mathbf{a} \times \mathbf{b} = \begin{pmatrix} 3 \\ 2 \\ 6 \end{pmatrix},\]then find the smallest possible angle between $\mathbf{a}$ and $\mathbf{b},$ in degrees. Answer: $30^\circ$}}\\
        \midrule
         \textbf{Orign}:\small{\texttt{Step 1: We know that for any two vectors $\mathbf{x}$ and $\mathbf{y},$ $\left(\mathbf{x} \times \mathbf{y}\right) \cdot \left(\mathbf{x} \times \mathbf{y}\right)=0.$This is because cross product of a vector with itself results into zero. Step 2: For our given problem, we have, \[(\mathbf{a} \times \mathbf{b}) \cdot (\mathbf{a} \times \mathbf{b})=\|\mathbf{a}\|^2\|\mathbf{b}\ |^2-((\mathbf{a} \cdot \mathbf{b}))^2.\] Since $(\mathbf{a} \times \mathbf{b}) \cdot (\mathbf{a} \times \mathbf{b})=33$, $\|\mathbf{a}\|=2$, and $\|\mathbf{b}\|=7$, then $(\mathbf{a} \cdot \mathbf{b})^2=(4)(49)-33.$Taking square root on both sides gives us $\mathbf{a} \cdot \mathbf{b}=5$or $-\mathbf{a} \cdot \mathbf{b}= -5.$We will now use this result to calculate minimum value of $\theta$ using dot product formula.  Step 3: Recall that the dot product of two vectors can be written as: \[\mathbf{a} \cdot \mathbf{b} = \|\mathbf{a}\| \| \mathbf{b}\| \cos{\theta}. \] Using the fact from above steps, we get: \[5 = (2)(7)\cos{\theta}\]\[5/14 = \cos{\theta}\] To solve for theta, take arccos of each side:\[\arccos{(5 / 14)} = {\theta}\] Convert theta from radians to degrees using radian conversion factor ($180/\pi)$: \({\theta} = {73^\circ }\) The final answer is: $\boxed{73}$}}\\
        \midrule
       \textbf{Steering}: \small{\texttt{(The vector cross product of two vectors is defined as follows: $\text{ Find} \mathbf{u} \times \mathbf{v}:$ - Take the first component of $ \mathbf{ u} $ and multiply it by the second component of $ \mathbf{ v}; $ - Multiply the second component of $ \mathbf{ u}$ by the third component of $ \mathbf{v} ; $ - Subtract the result from Step 1 from the result from Step 2; )  Step 1: Recall the formula for calculating the magnitude of a vector. To calculate the magnitude, or length, of a vector $\mathbf{x}$ with components $(x_1,x_2,...,x_n)$ we use the formula $\left | x \right |=\sqrt{{x}_{1}^{2}+{x}_{2}^{2}+\cdots +{x}_{n}^{2}}$. However, this problem does not require us to directly apply this formula.  Step 2: Use the given information about the magnitudes of vectors $\mathbf{a}$ and $\mathbf{b}$ along with their cross product to set up an equation involving the cosine of the angle between them using the relationship derived from the definition of the dot product and properties of orthogonal unit vectors (${\hat{i}}, {\hat{j}}, {\hat{k}}$). Given the cross product $\mathbf{a} \times \mathbf{b} = \begin{pmatrix} 3 \\ 2 \\ 6 \end{pmatrix}$ and knowing its magnitude can be used to relate to the angles between $\mathbf{a}$ and $\mathbf{b}$, let's proceed to derive this relationship.  Step 3: Calculate the magnitude of the cross product $\mathbf{a} \times \mathbf{b}$ which will allow us to determine the sine of the angle between vectors $\mathbf{a}$ and $\mathbf{b}$. We have $\|\mathbf{a} \times \mathbf{b}\|=\sqrt{(3)^2+(2)^2+(6)^2}=7.$  Step 4: Apply the relation between the norms of vectors $\mathbf{a}$ and $\mathbf{b}$, the norm of their cross product, and the sine of the angle between them based on trigonometric identities related to right triangles formed when you take the cross product of two vectors. Since $\sin(\theta) = \frac{\|\mathbf{a} \times \mathbf{b}\|}{\|\mathbf{a}\| \cdot \|\mathbf{b}\|},$ then substituting our known values gives $\sin(\theta)=\frac{7}{(2)(7)}=0.5.$ We need to solve for theta next.  Step 5: Solve for the angle $\theta$ by taking the arcsine of both sides after isolating sin($\theta$), remembering to convert radians to degrees if necessary since the solution asks specifically for degrees. Solving for $\theta$: $$\theta=\arcsin(0.5).$$  Step 6: Convert the value obtained in step 5 into degrees because the problem specifies that the answer should be provided in degrees. Using knowledge of common angles: $\arcsin(0.5) = 30^{\circ}.$ Therefore, the smallest possible angle between vectors $\mathbf{a}$ and $\mathbf{b}$ is $30^\circ$. The final answer is: $\boxed{30}$}}\\
        \bottomrule
    \end{tabular}
    \caption{Steering Case. LLM: Llama3.1-8B-it Featue:15593(numeric expressions and mathematical components) }
\end{table*}

\end{document}